# Judge AI: A Case-Study of Large Language Models as Judges

**Eric A. Posner[1] and Shivam Saran[1]**

**Abstract**

Can large language models (LLMs) replace human judges? By replicating a prior 2 × 2 factorial experiment conducted on 31 U.S. federal judges, we evaluate the judicial ability of OpenAI's GPT-4o. The experiment involves a simulated appeal in an international war crimes case, with two altered variables: the degree to which the defendant is sympathetically portrayed and the consistency of the lower court's decision with precedent. We find that GPT-4o is a competent judge who applies precedent correctly. GPT-4o disregards the illegally irrelevant factor of sympathy, similar to students who were subjects in the same experiment but the opposite of the professional judges, who were influenced by sympathy.

> "I predict that human judges will be around for a while." – Chief Justice John G. Roberts, Jr. (2023)

## 1. Introduction

Will AI ever replace human judges? Chief Justice Roberts suggests not anytime soon. We explore the question by instructing a popular LLM to "decide"[1] cases under experimental conditions nearly identical to those in which a group of human judges decided the same cases. We find that the LLM competently decides cases by correctly applying precedent. But, ironically, the LLM performs differently from the human judges by following the law more accurately than the judges did. But does that mean the LLM was a better judge—or a worse judge? The LLM's formalism closely matched that of a group of student subjects who, most would agree, were not qualified to be judges. In this paper, we describe our method and results, and conclude that the answer to our question may depend less on AI's progress than on jurisprudential puzzles that have stumped scholars for centuries.

This paper repeats experiments in human judicial decision-making reported in earlier papers by Spamann and Klöhn (2016, 2024) using an LLM rather than human subjects. We use these studies for two reasons. First, they are methodologically state-of-the-art and the authors made available data and other resources that facilitate replication of their results. This enabled us to maintain consistency in approach across their experiments and ours. And, unlike published case law, which is often used in judicial behavior studies (see Part 1 below), the experimental setup allowed for controlled variation and provided all the materials that judges use when they decide cases. Published cases do not include the factual record and usually not all the briefing. Published cases, which often run for many pages, also contain all kinds of details—many of them irrelevant—that would add noise to results and interfere with comparisons. And studies based on published opinions are subject to selection effects, as the vast majority of cases are settled or otherwise resolved without a published opinion. Published cases also introduce a specific challenge for LLMs: contamination risk, where the model may have already been exposed to the case (including its outcome) in its training data, thus influencing the model's decision. While this approach might be useful, we argue that a controlled

[1]University of Chicago Law School, Chicago, IL, USA

**Corresponding Author:**
Eric A. Posner, University of Chicago Law School, 1111 E 60th St, Chicago, IL 60637, USA.
Email: eposner@uchicago.edu

experiment is the best course of action at this earlier stage of research.

Second, because these experiments were conducted not only on human judges but law students as well, they provided us with a way of thinking about how the LLM behaves. Both judges and law students have legal training (some legal training, in the latter case), but judges also have a great deal of experience as lawyers and judges. We would expect students to decide cases in a more mechanical or legalistic way than judges do, and it will be useful to see whether LLMs decides cases more like students or more like judges.

Spamann and Klöhn's (2016) experiment was based on a real-world case before the International Criminal Tribunal for the Former Yugoslavia. The experiment closely replicated the structure and facts of the actual case with slight modifications along two dimensions: the degree of sympathy of the defendant and the strength of the relevant precedent. A group of 31 U.S. federal judges were instructed to decide the case individually. The judges, on average, discounted precedent and were more likely to rule in favor of sympathetic defendants than unsympathetic defendants despite the absence of legal relevance of degree of sympathy. In a follow-up paper, Spamann and Klöhn (2024) repeated the experiment using students as subjects, and found that the students were more likely than the judges to follow precedent and less likely to be influenced by the sympathetic nature of the defendant. This style of decision-making is known as "formalism" among legal scholars; the judges' style, which disregards the law when it leads to results judges think are wrong as a matter of ideology, personal preference, public policy, or common morality, is known as "legal realism" or just "realism."

Spamann and Klöhn (2016) argued that the experiments were consistent with a legal realist view of judges: judges are less influenced by the law than formalism would require. Sympathetic aspects of a party are rarely relevant to resolutions of legal disputes, and are not relevant at all in the war crimes case used by Spamann and Klöhn. And yet both common experience and empirical research do indicate that judges may be influenced by legally irrelevant factors including sympathy (Katz, forthcoming). The authors add that, for just this reason, students should not be used in studies of judicial decision-making. As we will discuss, in our initial experiments using an older version of ChatGPT (GPT-4o), which was the most advanced LLM at the time, the LLM behaved more like the students than like the judges: the LLM is formalistic. A robustness test considered numerous other LLMs and largely confirmed our results for GPT-4o. We limit ourselves to off-the-shelf models because they are already being used by judges, arbitrators, and organizations, and, as far as we know, no specialized model has been developed to serve a judicial function.

We offer three takeaways. First, off-the-shelf LLMs perform competently in this judicial experiment. They decide cases (and provide brief judicial opinions) in a way that any lawyer would see as coherent and familiar. Second, while the LLMs decide cases similar to human judges, they deviate in one key respect—they disregard the legally irrelevant factor of sympathy while human judges are influenced by it. Third, in this respect the LLMs decide cases more like student subjects than like professional judges. In broad terms, the LLMs and students are more formalistic, while the human judges are legal realists. We will discuss a number of possible explanations for these results, and their potential implications for LLM-based judging in the future as LLMs continue to develop.

This paper proceeds as follows. In Part 1, we provide background on the burgeoning law and AI literature and the mature judicial behavior literature. In Part 2, we describe our methodological approach. Part 3 provides our results, and Part 4 reports our attempts at prompting GPT to emulate federal judges. Part 5 discusses our findings.

## 2. Literature Review

### 2.1. AI and Law

The development of artificial intelligence (AI) has produced a host of novel applications to the law, from predicting court outcomes (Alghazzawi et al., 2022; Medvedeva et al., 2020; Shaikh et al., 2020) and detecting financial fraud (Ali et al., 2022; Gandhar et al., 2024; Sadgali et al., 2019; Hernandez Aros et al., 2024) to automating contract review. Since legal analysis is largely text-based, LLMs provide a unique vantage point. Their ability to process vast amounts of information enables them to quickly analyze complex legal documents, while their generative AI capabilities allow for nuanced decision-making and the production of legal content. One scholarly initiative, LegalBench, has demonstrated that LLMs can perform a wide range of legal tasks—162 tasks across six broad legal reasoning categories—with varying levels of effectiveness depending on the model and task at hand (Guha et al., 2023).

The most relevant strand of literature for our paper addresses the use of LLMs for legal reasoning. Several studies have claimed that LLMs have a promising ability to interpret legal language. Al Zubaer et al. (2023) found that domain-specific models, such as Legal-BERT and RoBERTa, exhibited strong performance in "argument mining" (for a discussion of mining, see Lawrence & Reed, 2020) within the European Court of Human Rights corpus, effectively identifying key argument components such as premises and conclusions. Choi (2024) tasked GPT with identifying a canon of construction known as the rule against surplusage in Supreme Court opinions. He found that GPT could not only pinpoint instances of its use with accuracy comparable to that of human research assistants but could also explain how such instances represented an application of the rule. Thalken et al. (2023) similarly examined LLMs' understanding of different jurisprudential methods employed in Supreme Court opinions, finding that fine-tuned models, such as Legal-BERT, are capable of distinguishing formal reasoning (i.e., reasoning strictly in accordance with laws) from "grand reasoning" (reasoning that considers other political, social, and economic factors).

Other work has investigated the capacity of LLMs to master narrow areas of the law. Nay et al. (2024) tested whether GPT could answer multiple-choice questions covering the U.S. tax code and Treasury regulations, and found that the best-performing model achieved roughly 70% and 90% accuracy, respectively. Hassani (2024) presented various LLMs—GPT, Mistral, and BERT—with food safety and data privacy regulations, and found that they could accurately classify the regulations' legal provisions into key compliance categories. This included independently assigning compliance-relevant labels, as well as sorting provisions into predefined categories (e.g., "Color" or "Pathogen" for food safety regulations). Kang et al. (2023) asked ChatGPT legal questions based on 50 hypothetical scenarios related to Malaysian contract law or Australian family law, and found that it could produce correct and reasonable answers, along with even better responses when supplemented with relevant legal context. Nelson (2023) found that ChatGPT competently interprets international treaties. Coan and Surden (2024) examined LLM's performance in constitutional interpretation, finding that, although sensitive to prompts, the models could efficiently summarize and extract key details from legal texts with notable accuracy. Nyarko et al. (2025) found that a group of law professors preferred an LLM's answers to office hours-style student questions about contract law over answers provided by other law professors.

Engel and McAdams (2024) tested GPT's ability to determine the ordinary meaning of statutory terms by presenting it with the well-known "no vehicles in the park" hypothetical. They asked GPT to determine whether various objects (e.g., bicycles, skateboards) qualify as "vehicles" and found that GPT can produce responses that align reasonably well with those of human respondents. Similarly, Arbel and Hoffman (2024) asked various LLMs—GPT, LLaMA, and Claude—to define ambiguous terms in legal contracts, and found that these models have the capacity to "generatively interpret" the intended meaning of particular terms at the time of the contract's writing. For example, when assessing a royalty agreement that was adjudicated by the New York Court of Appeals in Ellington v. EMI Music, Inc. (21 N.E.3d 1000 [2014]), the LLMs interpreted the phrase "other affiliates" to include not only affiliates at the time of the agreement but also those that may have arisen over time. This distinction was a key point of contention in the case, yet one the court overlooked. The authors also found that the LLMs can aid in quantifying the level of ambiguity in specific terms. By converting different words and phrases—in this case, the different possible interpretations of the term—into high-dimensional numerical representations known as embeddings, they measured how similar or dissimilar each interpretation was to the original term (for a discussion of embeddings, see Bingi & Yin, 2024).[2] For instance, in assessing what the term "flood" means in an insurance contract, the LLMs found "heavy rainfall" to be more numerically fitting than "burst pipe."

Beyond their ability to interpret legal language, LLMs may be able to assist with legal decision-making (Dhungel, 2024). For instance, He et al. (2024) reports that LLMs can simulate court debates that mirror real-world courtroom interactions, retrieve and generate relevant legal articles and precedents, and even issue judgments. Nay (2023) found that LLMs can adequately interpret fiduciary obligations, reporting that GPT was able to correctly predict court outcomes in cases involving breaches of fiduciary obligations with high accuracy. Moreover, he found that GPT's performance has gradually improved over time, with GPT 3.5 (the most recent model at the time of publication) achieving 78% accuracy, compared to OpenAI's previous models GPT-3 and Curie achieving 73% and 27% accuracy, respectively. Shui et al. (2023) presented various LLMs—GPT, LLaMA, BLOOMZ, and ChatGLM—with a list of case facts, alongside multiple-choice options for various charges, and found that the LLMs could accurately predict the appropriate charge. Menezes-Neto and Clementino (2022) examined several advanced LLMs, including ULMFiT, BERT, and Big Bird, for their ability to predict outcomes in Brazilian federal court appeals and found that they outperform human experts. Engel (2025) used LLMs in an experiment to determine whether varied contract terms impact supplier profits and enable price discrimination.

Lastly, several studies have explored the practical applications of LLMs in the legal context. Chien et al. (forthcoming) finds that LLMs can be useful alternatives for low-income individuals who otherwise would not have the resources for legal services. Utilizing OpenAI's custom GPT feature, they developed two chatbots tailored to Arizona law that provides low-income Arizonans with relevant legal advice: one for marijuana cases and another for evictions. Both Choi and Schwarcz (forthcoming), and Choi et al. (forthcoming) found that GPT can significantly enhance exam performance for lower-performing law students. Cope et al. (2025) found that LLMs are comparable to law professors at grading law exams. Using exams from top-30 U.S. law schools, LLM grades correlate with professor grades at Pearson coefficients of up to 0.93.

While these studies demonstrate the potential of LLMs, they all face significant limitations. In many studies, it is unclear what authors mean when they argue that the LLM performs "competently" or "reasonably." In studies that compare LLM performance to that of human research assistants, it is unclear what kind of baseline these assistants or student subjects provide, compared to real legal decision-makers who are experienced professionals. Many of the studies also confront familiar problems with LLM hallucinations (output of false or nonsensical information). Dahl et al. (2024), for instance, report that LLMs hallucinate between 58% and 88% of the time when faced with legal queries and tasks. Certain precautions, such as proper prompt engineering and selecting the best interaction techniques,[3] may mitigate the risks of LLM hallucinations. But their capacity to reduce errors to an acceptable level remains unknown (Homoki & Ződi, 2024; Lai et al., 2023). The rapidly growing literature also develops methods for testing LLMs; in addition to the citations above, see Engel, (2025), Engel et al., (2025).

### 2.2. Judicial Behavior Literature

The judicial behavior literature is a vast, decades-long project of using statistical and experimental methods to understand how judges decide cases. The literature can be traced back to the legal realist movement of the early 20th century. The legal realists argued that traditional legal scholarship, which assumed that judges decide cases by applying established legal rules to novel facts, misrepresented the reality of judicial decision-making. Realists believed that judges are influenced by extralegal policy considerations or psychological quirks. The modern judicial behavior literature explores these and related hypotheses (for example, that judges are influenced by career concerns) by using statistical analysis of judicial decisions, votes, and related behavior (for a survey, see Epstein, 2016). Early on, scholars established that the voting of U.S. Supreme Court justices partly reflected their ideological views, as proxied by the party affiliation of the president who nominated them, the public attitude toward them as shown by newspaper articles, and so on (see Baum, 1997, 2008). The literature has since branched in many directions. It has examined judicial behavior at all levels and in foreign countries; the impact of panel composition on the decisions of appellate bodies; the impact (or lack of impact) of specific rules that attempt to control judicial discretion, like the *Chevron* doctrine; the impact of elections on the decisions of elected judges; and much else (see Epstein, 2016).

Despite all this work, judicial decision-making remains poorly understood. While the literature has made significant progress in understanding which extralegal factors influence judges, it has made little progress in understanding the magnitude of the effect. One source of frustration is the methodological limits of relying on reported decisions, which reflect selection effects. And because the facts of every case differ, it is impossible to rule out omitted variables in regressions. These limitations have led some scholars to conduct experiments using real judges or (because real judges are rarely willing to sit for a study) student subjects (for some examples, see Guthrie et al., 2001; Wistrich et al., 2015; Rachlinski & Wistrich, 2019, 2021; Katz, forthcoming; Klerman & Spamann, 2024; and see Holste & Spamann, forthcoming).

This work was the inspiration for our study of the judicial potential of LLMs. Because LLMs have not been used as judges, we cannot study their judicial output by regressing decisions or votes on independent variables of interest as the judicial behavior literature does. But we can subject LLMs to the same experiments that have been conducted on human judges. Our initial thinking was that if LLMs perform like human judges in the experiments, that could be the basis for optimism that, contrary to Justice Roberts' prediction, LLMs could serve as cheap, accurate, and tireless adjudicators who do not suffer from ideological biases, careerist instincts, and other human limitations.

## 3. Methodology

### 3.1. Spamann and Klöhn Experimental Design

Spamann and Klöhn (2016) based their experiment on a case that came before the International Criminal Tribunal for the Former Yugoslavia (ICTY), *Prosecutor v. Momčilo Perišić* (case no. IT-04-81). Perišić was charged under Article 7 of the ICTY Statute with "aiding and abetting the planning, preparation, or execution of crimes" against Muslim civilians during the civil war in Yugoslavia. In his role as Chief of the General Staff of the Army of Yugoslavia (VJ)—the VJ's highest-ranking position—Perišić provided substantial support to the Army of Republika Srpska (VRS), the primary military force of the Serbs at the time. The prosecution alleged that, through his support, Perišić facilitated the VRS's operations, which systematically targeted Muslim enclaves, resulting in serious injuries and deaths among civilians. While the trial chamber found Perišić guilty (case no. IT-04-81-T, 2011), the Appeals Chamber overturned his conviction (case no. IT-04-81-T, 2013), ruling that the prosecution had not proven that Perišić's support was "specifically directed" toward the criminal activities of the VRS, a requirement under Article 7(1) of the Statute. The Appeals Chamber held that the evidence did not "establish a sufficient link between aid provided by an accused aider and abettor [Perišić] and the commission of crimes by principal perpetrators [VRS]."

While the case involved a number of legal issues, Spamann and Klöhn (2016) focused on the concept of "specific direction," the central issue in the case. The subjects were professional judges who were asked to assume the role of an Appeals Chamber judge and decide whether to affirm or reverse the trial chamber's decision.[4] The opinion of the lower court was provided to the judges almost exactly as it was originally written, with a few variations described below. Judges were also provided with a statement of agreed facts, briefs from the prosecution and the defense, the ICTY statute, and a former ICTY Appeals Chamber decision to serve as precedent.

To determine which factors influenced judges' decisions, Spamann and Klöhn (2016) conducted a 2 × 2 factorial experiment in which two elements were varied: sympathy and precedent. To vary sympathy, Spamann and Klöhn (2016) presented the defendant as either sympathetic or unsympathetic (that is, in the sense of deserving of sympathy). To do this, they replaced the original defendant, Perišić, with two fictitious profiles. The first, Ante Horvat, was described as a sympathetic Croat who expressed deep regret for the conflict's violence and, post-war, became vice-chairman of the Croatian-Bosnian Reconciliation Commission. The second, Borislav Vuković, was described as an unsympathetic Serb who publicly ridiculed the tribunal and showed no remorse for his actions. The details that depicted a defendant's level of sympathy were included only in the agreed facts and the briefs, which were written by Spamann and Klöhn. The trial case presented to the judges was nearly identical to the original Perišić case, with only the name of the defendant (and for Horvat, his nationality) changed.

To vary precedent, the study exposed half of the judges to a precedent that directed them to affirm the conviction (i.e., "specific direction" is not a necessary element in aiding and abetting liability), and the other half to a precedent that suggested reversal (i.e., establishing "specific direction" is necessary for criminal liability). These precedents were real decisions by the ICTY Appeals Chamber, *Prosecutor v. Nikola Šainović* et al. (case no. IT-05-87-A, 2014) (holding that "specific direction" is a not a necessary element) and *Prosecutor v. Mitar Vasiljević* (case no. IT-98-32-A, 2004) (holding that it is). Table 1 provides the matrix.

The hypothesis that realists care about sympathy and formalists do not generates conflicting outcomes in two of the four variations: when precedent requires affirmance of the conviction of a sympathetic defendant (box 2) and when precedent requires reversal of an unsympathetic defendant (box 3). When sympathy and precedent point in the same direction, realists and formalists should agree (boxes 1 and 4). Note that neither precedent nor sympathy necessarily dictates a single result: a subject may be influenced by both precedent and sympathy but weigh them differently. Thus, the outcomes in Table 1 should be understood in relative rather than absolute terms. A realist is more likely to reverse in box (2) than in box (1) but may affirm (or reverse) in both. Realism and formalism are a matter of degree.

It is important to acknowledge that the experimental setup treats the precedent as "persuasive" rather than "binding." In law, that means that judges are not required to follow the precedent, and so the experimental setup would not compel a formalist to follow the precedent. But judges are allowed to, and in practice likely to, follow persuasive precedent even if they might interpret the statute otherwise. There are several reasons for this. The earlier decision and reasoning provide an independent source of information (unlike the briefing); moreover, consistency has value both for the legitimacy of the judicial system and for fairness of regulated parties. Because precedent is a legally relevant variable and sympathy is not, formalist judges should follow precedent more often than sympathy, though they will not necessarily always follow precedent.

Spamann and Klöhn's experiment suffers from a few limitations, the most obvious of which is the small sample size. This may account for one anomalous result. As we discuss below, the authors find that more judges affirm the conviction of a sympathetic defendant when precedent dictates reversal than when the precedent dictates conviction. Neither students nor GPT produce a similar anomaly. It is also not clear that the level of defendant's sympathy is legally irrelevant as the authors posit, as a judge could arguably think that repentance or remorse implies that the defendant lacked mens rea.[5]

## 3.2. Adapted Experimental Design

To replicate Spamann and Klöhn's analysis, we first compiled all the material provided to judges in their original experiment. These materials included: (1) instructions detailing the judge's role, (2) a statement of the agreed facts, (3) briefs for the prosecution and the defendant, (4) the ICTY statute, (5) a former appeals chamber decision to serve as precedent, and (6) the lower court's trial judgment.

To adapt these materials for GPT, we made several modifications. First, the instructions were altered to suit the format and constraints of presenting information to a large language model, as opposed to human judges.[6] We removed certain elements that were irrelevant or impractical in the LLM context (e.g., time constraints) and added clarifications to facilitate the LLM's comprehension of its processing capabilities (i.e., token limits[7]).

We also made some other adjustments to the experiment. The statute, precedent, and trial judgment were originally presented to judges as hyperlinked documents on computers used in the experiment. Judges were able to navigate through these documents, one tab at a time, opening and closing as needed. Since GPT is incapable of opening links and browsing their content, we converted these materials from html documents to plain text.[8]

The statute, once in text format, was given to GPT verbatim as it was given to the human judges. The trial judgments and precedents, however, being complex cases, ran quite long, averaging roughly 229,000 tokens (164,000 words) and 54,000 tokens (37,000 words), respectively. Because of GPT-4o's token limit (128,000 tokens), it was infeasible to provide these documents in their entirety alongside the other material.

**Table 1.** Hypothesis Matrix

| | Precedent: Affirm conviction | Precedent: Reverse conviction |
|---|---|---|
| Unsympathetic defendant | (1) Affirm | (3) Formalist: Reverse<br>Realist: Affirm |
| Sympathetic defendant | (2) Formalist: Affirm<br>Realist: Reverse | (4) Reverse |

Moreover, the GPT API currently does not provide a feature allowing the model to retain memory.[9] This means we could not feed GPT one document at a time, expecting it to retain past information, unlike the online interface for ChatGPT. To overcome this challenge, we instead instructed GPT to create summaries of each case, extracting key elements and any relevant information for its legal understanding.[10] This technique has both advantages and drawbacks (see Liu et al., 2023). Beyond circumventing the token limit, summarizing allows for the extraction of the most substantive material, stripping away extraneous details. However, such loss of detail may also result in the omission of nuances that are critical for a full legal analysis.

The statement of facts and briefs for both sides were presented to GPT verbatim, as these documents required no modifications. Thus, our final materials included: (1) LLM-specific instructions, (2) the unaltered statement of agreed facts, (3) unaltered briefs for the prosecution and defendant, (4) the unaltered statute, (5) the summarized precedent, and (6) the summarized trial judgment. We then instructed GPT to determine whether the lower court's decision should be affirmed or reversed based on the information presented to it, and to provide a brief paragraph describing its rationale.[11]

### 3.3. GPT Specifications

We used GPT-4o,[12] the latest release of GPT at the time we began this project.[13] One advantage of GPT-4o is its high token limit of 128,000 tokens, which is considerably greater than OpenAI's previous flagship model, GPT-3.5-turbo (16,385 tokens), as well as most other available LLMs at the time of experimentation.[14] This higher limit allowed us to provide the lengthy trial judgments and precedents, although, as mentioned earlier, it was insufficient to process all of the input material in a single pass.

The only parameter that we specified was the temperature value, which controls the degree of randomness in an LLM's output—that is, how likely the model is to deviate from the most probable outputs. GPT's temperature parameter ranges from 0 to 2, with higher values representing more randomness. The default value is set to 1.

Selecting the right temperature depends on the task at hand. For tasks requiring consistent, repeatable results, such as testing the model's understanding of a particular legal concept or its ability to classify legal documents, the temperature is often set to 0 (see, e.g., Nay et al., 2024; Choi, forthcoming). On the other hand, more interpretative tasks, such as determining the meaning of statutory terms or contracts, typically use temperatures near 1 to allow for a greater variety of responses (see, e.g., Arbel & Hoffman, 2024; Engel & McAdams, 2024). In our experiment, we sought a middle ground that reflected the natural variability in judicial decision-making. A temperature too low would likely produce the same result each time (e.g., always affirm), which would fail to capture the diversity seen in real-world judicial outcomes. Conversely, with a temperature too high, GPT would return nonsensical decisions, rendering the results useless. Based on insights from others in the related academic literature (see, e.g., Engel & McAdams, 2024, p. 15), as well as the broader consensus among AI experts (Chang et al., 2023; Peeperkorn et al., 2024; Renze & Guven, 2024; Zhu et al., 2023), we determined that temperature values above 1 tend to introduce too much noise. Thus, we set the temperature to 0.7, a commonly selected setting in scholarly work aimed at balancing coherence and creativity (see, e.g., Mukherjee et al., 2023; Abramski et al., 2023; Osmanovic-Thunström & Steingrimsson, 2023; see broadly Ramlochan, 2024). This allowed for some variation in responses without straying too far from the logical consistency expected of judicial rulings. As a robustness test, we reran the experiment with temperature values of 1 (the default) and 0.3, and found no significant differences in our results. Full robustness results are reported below. As OpenAI recommends when altering temperature values, we left the top_p parameter untouched.[15]

We assigned GPT the following system prompt: "You are an appeals judge in a pending case at the International Criminal Tribunal for the Former Yugoslavia (ICTY). Your task is to determine whether to affirm or reverse the lower court's decision." System prompts help set the context for LLMs, informing them of the role they should play and the instructions they need to follow. These prompts precede and function independently of the user input (in our case, the various materials). For example, if the system prompt was set to "Respond in all caps," then the LLM would respond in uppercase letters, even if the user prompt did not so specify. While system prompts are especially useful in back-and-forth conversations, where specific instructions for the model may need to be maintained, they are also helpful in single-call interactions (as in our study), as they help the model establish the necessary context before processing the user input.

Whereas Spamann and Klöhn's experiments involved 31 judges and 91 students, ours used only a single large language model: GPT 4o. In order to artificially increase our sample size, we used different seed numbers. The seed feature allows a GPT model to have a specific computer formulation by locking in the sequence of random decisions it makes and associating it with a distinct reference number (e.g., 1294802). Therefore, API calls with the same parameters, prompt, and seed number should provide the same output.[16] Keeping the seed number consistent while varying the prompt allows us to isolate the effect of changing the prompt, or in our case, the condition. In total, we generated 25 random seeds, used on each of the four conditions (Sympathetic/P-Affirm, Sympathetic/P-Reverse, Unsympathetic/P-Affirm, and Unsympathetic/P-Reverse), yielding a total sample of n = 100.

Increasing the sample size is necessary for several reasons. First, the large sample size enables us to analyze GPT's results in aggregate. Since individual outputs from GPT can vary, even with the same prompt, a larger sample allows us to determine the "average" GPT response—in this case, the rate of affirmance. This serves as a robustness check against the inherent stochasticity of LLMs. Second, the large sample size increases statistical power. Note, however, that the comparison between GPT and human subjects is not apples to apples. For GPT, regression results and p-values should be interpreted as describing the stability and conditional responsiveness of a single model under stochastic perturbation, rather than as population-level inference over independent decisionmakers. Variation in human judgments reflect life experiences, cognitive response in the experimental setting, and related factors. Variation across GPT seeds is purely stochastic. But the comparison is still meaningful. In choosing between human and algorithmic judges, we should take account the variance as well as the quality of the outcomes they generate.[17]

Third, a large sample size helps uncover patterns in the LLM's responses. Finally, it allows us to assess the impact of different prompt engineering techniques. With more trials, we can determine whether changes in prompts genuinely influence GPT's output, whereas in a smaller sample, any observed variation may result from model randomness. The difference in sample size between students, judges, and the LLM is significant. As a result, simple comparisons of outcome proportions will necessarily yield wider confidence intervals for judges than for GPT and students. For this reason, we present the initial comparisons for descriptive context and rely on regression analysis to more rigorously account for differences in sample size.

One might wonder why we did not fine-tune the model to better approximate human judicial behavior. There are three reasons. First, as discussed below, although human judges exhibited a sympathy effect, their written rationales never once actually mention sympathy; they rely on standard legal reasoning. Thus, fine-tuning with published opinions would be unlikely to teach the model to consider sympathy itself or other factors (like ideology) that affect decisions but are not explicitly incorporated into the text of the opinion. Any shift in decision-making that suggests increased resemblance to human outcomes would more plausibly reflect learned statistical associations between the case condition and the judges' decisions. Second, the experiment we study is highly specific (an international war crimes case), and thus poorly suited for fine-tuning. Even if fine-tuning did improve performance in this particular case, there would be no reason for us to believe such improvements would generalize to other contexts. Third, fine-tuning is not an easy or straightforward process. To fine-tune an LLM, one must make numerous contestable decisions. Among other things, one must determine the scope of the training data, the base model to be used, the degree and kind of intervention by experts, and the criteria used to evaluate outputs. It is not the purpose of this paper to determine whether it is theoretically possible for an LLM to resolve disputes in some kind of optimal fashion, though we believe that our results do provide some grounds for skepticism.

## 4. Results

### *4.1. Frequency of Affirming*

In our initial test, we generate LLM results for the four scenarios. Figure 1 shows the frequency of affirming across individual factor levels, with 95% bootstrap confidence intervals. Each bar reports the overall affirmance rate for a given factor level, pooling across the other factor (e.g., the first bar reports the affirmance rate when the defendant is sympathetic, regardless of whether the precedent advises affirmance or reversal). Figure 1 reveals two patterns. First, GPT's rate of affirmance is largely unaffected by whether the defendant is portrayed as sympathetic or unsympathetic. Second, GPT is more likely to affirm when the precedent supports affirmance and less likely to affirm when precedent supports reversal. To determine whether the proportion differences under the various conditions were statistically significant, we computed p-values using the Boschloo two-sided exact test.[18] We find that the difference in affirmance rates when the precedent says to affirm versus when the precedent says to reverse is statistically significant ($p < 0.01$), but the difference in affirmance rates for a sympathetic and unsympathetic defendant is not statistically significant. These results provide reason to believe that GPT is potentially a competent jurist. It follows the law and disregards a nonlegal factor. It certainly does not act randomly.

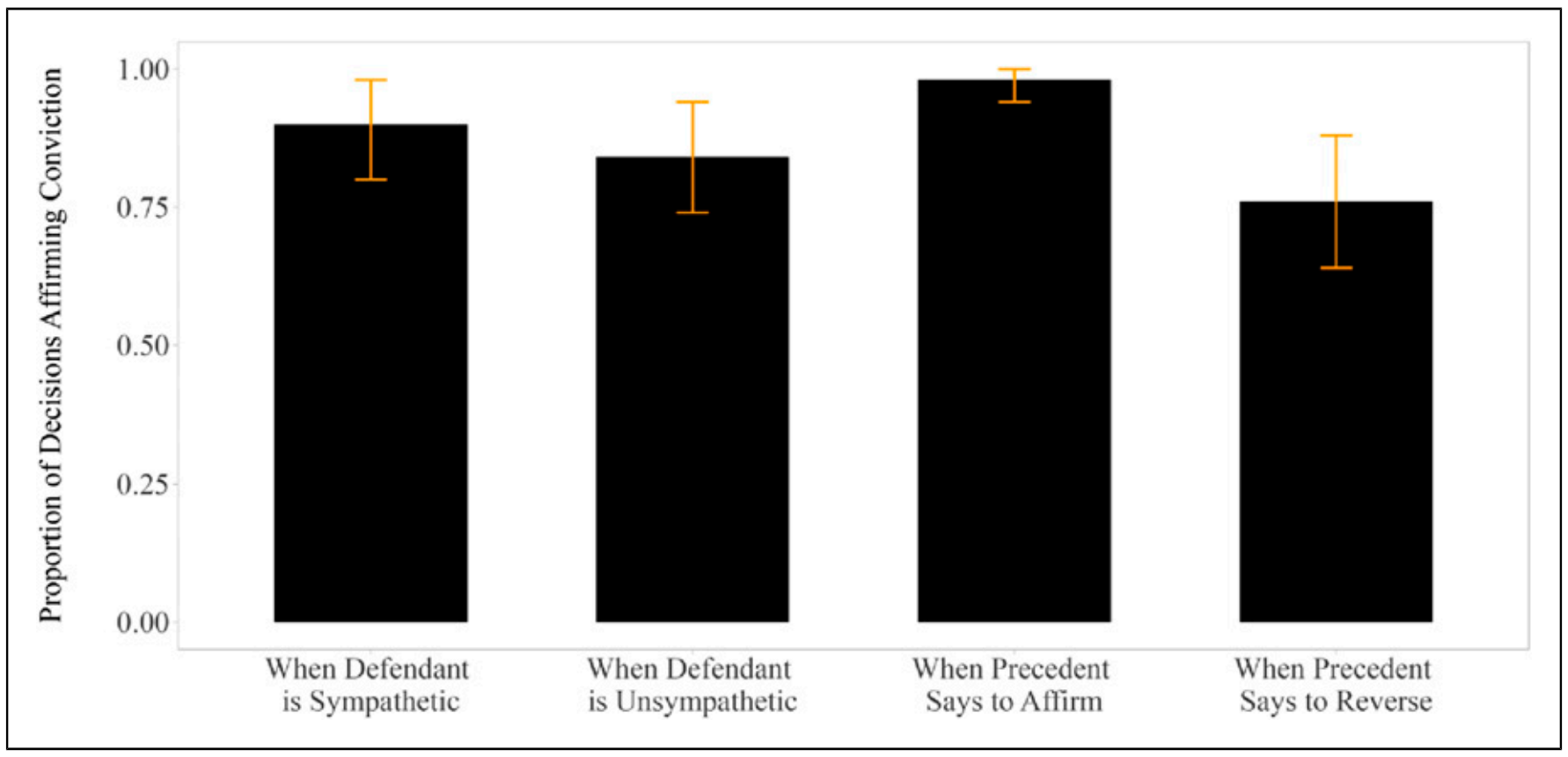

**Figure 1.** Frequency of affirming convictions: GPT results

In Figure 2, we compare the GPT results with those in the judge and student experiments. The first bar in each graph is the same as in Figure 1; the remaining two bars show affirmance rates for judges and students.

Here, a paradox emerges. While GPT consistently applies the law, the human judges do not. GPT performs more similarly to students than to judges. When precedent requires affirmance of a conviction (column 3), GPT and students are more likely to affirm than judges are (at a statistically significant level). When precedent requires reversal of a conviction (column 4), GPT is more likely to reverse than judges but not at statistically significant level. (As in Spamann and Klöhn's study, students are more likely to reverse than judges, though not quite at a statistically significant level ($p = .09$.) The level of defendant sympathy does not affect GPT's and students' decision whether to affirm, while judges are more likely to affirm the conviction of an unsympathetic defendant (column 2) than the conviction of a sympathetic defendant (column 1).

The alignment between GPT's behavior and that of students can largely be attributed to their shared formalistic approach—both are inclined to follow precedent, though GPT to a slightly lesser extent.[19] Compared to students, GPT shows a stronger tendency to affirm even when the precedent advises reversal, suggesting there may be an affirmance bias.[20] Nonetheless, both students and GPT prioritize precedent over sympathy. This pattern contrasts with that of judges, who give less weight to precedent. In fact, in Spamann and Klöhn's experiment, judges often did the opposite of what precedent suggested: they frequently affirmed when precedent advised reversal and reversed when it advised affirmance. To better understand these results, Figure 3 provides an analysis of the cross-interactions.

Let us begin by addressing the cases where sympathy and precedent indicate the same result. When precedent directs affirmance of a conviction and the defendant is unsympathetic, we expect all groups to affirm. That is the result, as shown in the third column of bars. When precedent directs reversal of a conviction and the defendant is sympathetic, we should expect subjects to be less likely to affirm. Indeed, that is the result in the second column, though interestingly students are affected more than GPT and the judges.

Where sympathy and precedent point in different directions, we get a better sense of the differences between the groups. When precedent directs affirmance of a conviction and the defendant is sympathetic, GPT and students are highly likely to affirm, while judges are more likely to reverse (column 1). When precedent directs reversal of a conviction and the defendant is unsympathetic, GPT and students are more likely to reverse than judges are (column 4).

In short, GPT and students are more formalist (in the sense of following precedent), and judges are less so (in the sense of being swayed by sympathy). The results are summarized in Table 2. As noted above, "affirm" and "reverse" refer to the relative likelihood of affirmance or reversal.

Table 3 provides the numerical results and levels of statistical significance for each of the conditions. The first column indicates the condition and the second indicates the comparison (either GPT vs. Judges or GPT vs. Students). The third column presents the raw difference in affirmance rates. These are calculated by subtracting the other group's (Judges or Students) rate of affirmance from GPT's rate of affirmance. Thus, a positive value indicates that GPT affirmed more than the other group, while a negative value indicates it affirmed less. Values approach zero as the two groups become more similar. To determine whether these differences were statistically significant, we computed p-values using the Boschloo exact test. These values are reported in column 4. Finally, column 5 provides Cohen's h effect sizes for a more nuanced metric of comparison.

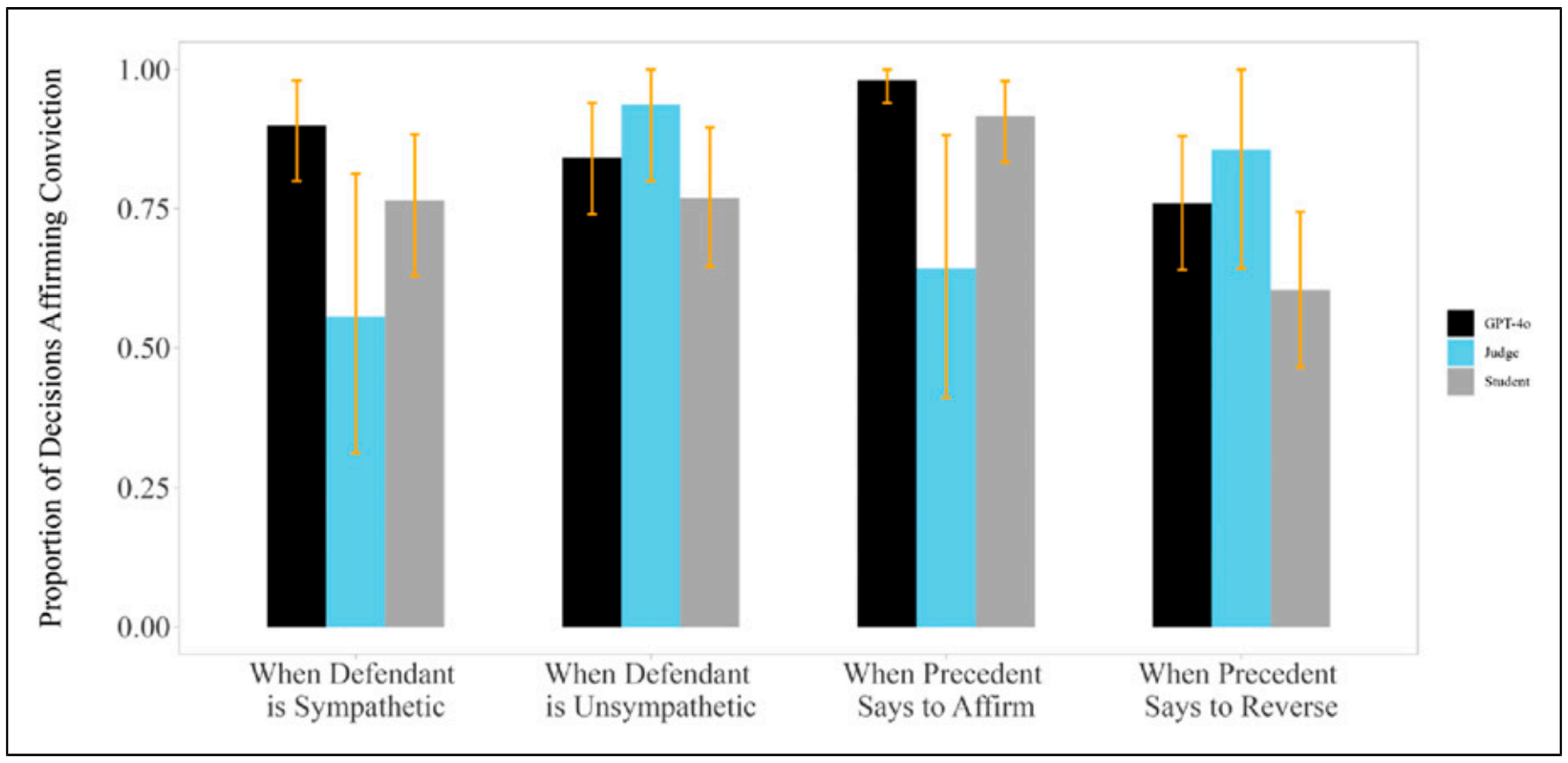

**Figure 2.** Frequency of affirming conviction: Comparison of GPT, judge, and student results

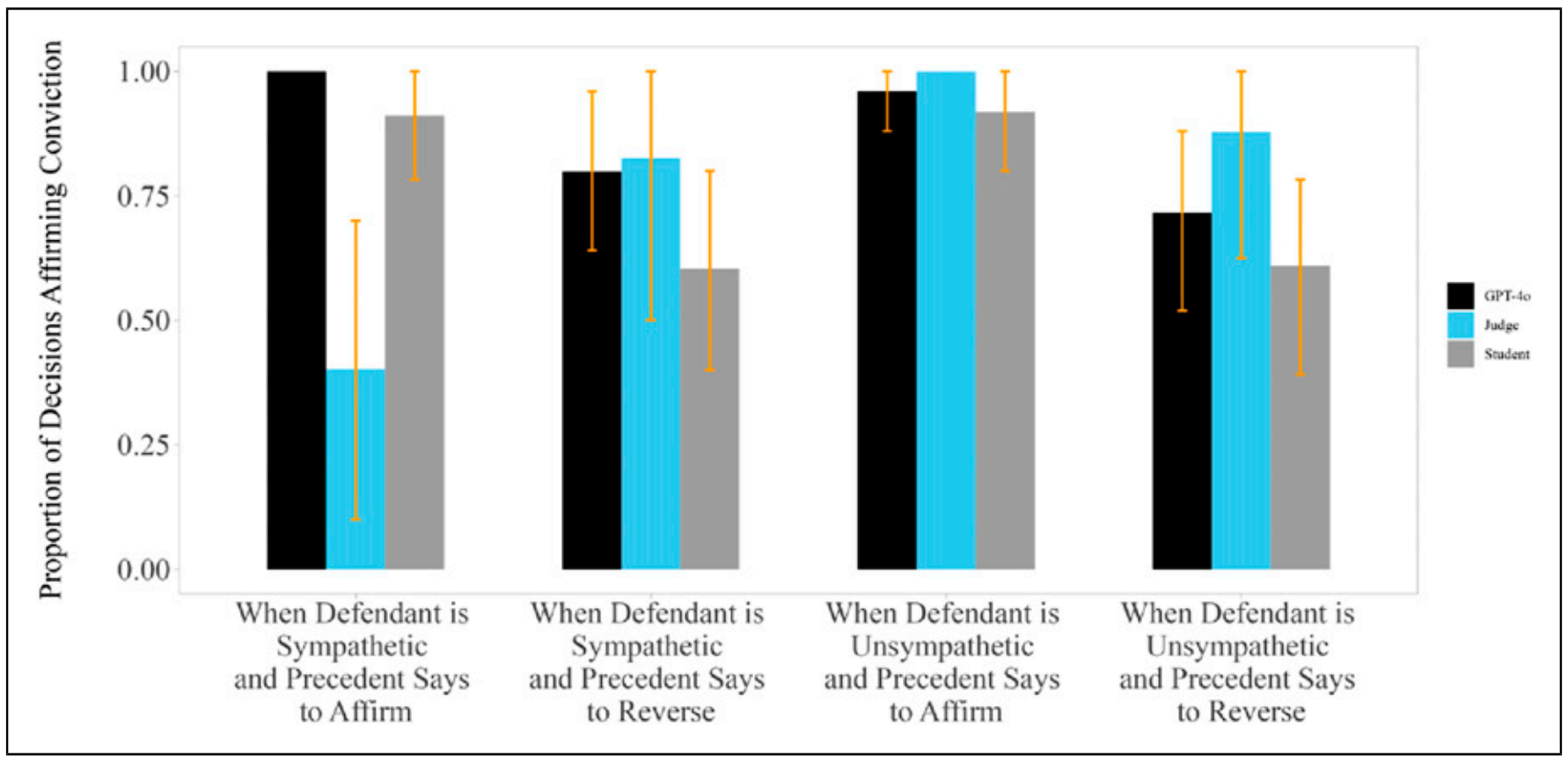

**Figure 3.** Comparative results by scenario

GPT performs closer to students than to judges in five of the conditions (P-Affirm, Sympathetic, Unsympathetic, Sympathetic/P-Affirm, and Unsympathetic/P-Reverse), closer to judges in two (P-Reverse and Sympathetic/P-Reverse), and equally distant from both in one (Unsympathetic/P-Affirm). The difference between GPT and judges is statistically significant under the conditions of P-Affirm, Sympathetic, and Sympathetic/P-Affirm. No statistically significant differences were found between GPT and students.

While proportion differences give a broad sense of similarity, they can sometimes oversimplify the comparison. For instance, while a 0.1 difference between proportions of 0.4 and 0.5 may seem small, the same 0.1 difference between proportions of 0 and 0.1 is more pronounced—it means that one group never affirms while the other affirms at least some of the time. To account for these subtleties, we also computed Cohen's h effect sizes. Cohen's h standardizes the difference between proportions, making it easier to assess how practically meaningful the difference is. A value of h = 0.2 represents a small difference, h = 0.5 a medium difference, and h = 0.8 a large difference. The Cohen's h values we calculated generally align with the proportion differences, but they offer additional insight in a few cases.

First, when the defendant is sympathetic and precedent advises affirmance, the proportion difference between GPT and students is only 0.09, yet the Cohen's h value of 0.6 indicates a medium effect size. This discrepancy arises because GPT affirms 100% of the time under this condition, while students do most, but not all, of the time. Second, under the Unsympathetic/P-Affirm condition, Cohen's h highlights a meaningful difference between GPT and judges (h = 0.4), even though the proportion difference is just 0.04. This is due to GPT affirming nearly every time (24 out of 25 instances) under the Unsympathetic/P-Affirm condition.

## *4.2. Regressions*

To ensure consistency with Spamann and Klöhn's methodology, we adopted the same regression models (and specifications) that they used to statistically test the difference in affirmance rates between GPT, judges, and students.[21] These models included an OLS model, a Logit model, and an Exact Logistic model. The baseline category was set with judges as the group, the defendant as sympathetic, and a precedent that supports affirmance. Table 4 reports the results, with OLS outcomes presented as linear regression coefficients, and the Logit and Exact Logistic outcomes presented as odds ratios; 95% confidence intervals are also reported.[22]

The results are consistent across all models and offer several insights. First, to reiterate Spamann and Klöhn's findings, human judges do not follow precedent at a statistically significant level and disfavor unsympathetic defendants at a statistically significant level.

**Table 2.** Comparative Results Matrix

| | Precedent: Affirm conviction | Precedent: Reverse conviction |
|---|---|---|
| Unsympathetic defendant | (1) Affirm [no conflict] | (3) GPT and Students: Reverse<br>Judges: Affirm |
| Sympathetic defendant | (2) GPT and Students: Affirm<br>Judges: Reverse | (4) Reverse [no conflict] |

**Table 3.** Comparing GPT's Frequency of Affirming to Judges and Students

| Condition | Comparison | Proportion difference | P-value | Cohen's h effect size |
|---|---|---|---|---|
| P-Affirm | GPT vs. Judges | 0.33 | <0.01 | 0.98 |
| | GPT vs. Students | 0.06 | 0.27 | 0.29 |
| P-Reverse | GPT vs. Judges | −0.10 | 0.50 | −0.26 |
| | GPT vs. Students | 0.16 | 0.12 | 0.35 |
| Sympathetic | GPT vs. Judges | 0.34 | 0.01 | 0.81 |
| | GPT vs. Students | 0.13 | 0.10 | 0.36 |
| Unsympathetic | GPT vs. Judges | −0.09 | 0.48 | −0.29 |
| | GPT vs. Students | 0.07 | 0.44 | 0.18 |
| Sympathetic/P-Affirm | GPT vs. Judges | 0.60 | <0.01 | 1.77 |
| | GPT vs. Students | 0.09 | 0.33 | 0.61 |
| Sympathetic/P-Reverse | GPT vs. Judges | −0.03 | >0.99 | −0.08 |
| | GPT vs. Students | 0.20 | 0.17 | 0.44 |
| Unsympathetic/P-Affirm | GPT vs. Judges | −0.04 | >0.99 | −0.40 |
| | GPT vs. Students | 0.04 | 0.89 | 0.17 |
| Unsympathetic/P-Reverse | GPT vs. Judges | −0.16 | 0.44 | −0.41 |
| | GPT vs. Students | 0.11 | 0.48 | 0.23 |

Second, while both students and GPT are more likely to affirm than judges are, the effect is stronger for GPT. The probability of an affirmance increases by 50.6% in the OLS model, and the odds increase by 78 times and 36 times in the Logit and Exact Logistic models, respectively, when GPT replaces a judge. In comparison, when a student is substituted for a judge, the probability of affirmance increases by 40.8% in the OLS model, and the odds increase by 12 times and 9 times in the Logit and Exact Logistic models, respectively. The smaller coefficients indicate that students perform closer to judges than GPT, though both act differently from judges at a statistically significant level.[23]

The interaction terms further clarify these patterns. The regressions confirm that GPT is much more likely than judges to reverse a conviction if precedent tells it to reverse, compared to when precedent says to affirm. The effect of a reversal precedent is estimated to be 37 percentage points stronger (OLS model), 50 times stronger (Logit), or 33 times stronger (Exact Logistic) for GPT than it is for judges. The opposite is true for sympathy; GPT is estimated to be 41 percentage points less likely (OLS model), 20 times less likely (Logit), or 14 times less likely (Exact Logistic) than judges to reverse a conviction when the defendant is unsympathetic, compared to when he is sympathetic. The total precedent effect for GPT is estimated to be −0.22 for the OLS model (0.15 + −.37) and 0.057 for the Logit model (2.85 * .02). That is, GPT is 22% or roughly 18 times less likely to affirm the conviction when precedent directs it to reverse, depending on the model. The total sympathy effect for GPT is −.06 for the OLS model (.35 + −.41) and 0.501 for the Logit model (10.15 * .05); that is,

**Table 4.** Regression Models

| | Dependent variable: Affirmed | | |
|---|---|---|---|
| | OLS | Logit | Exact logistic |
| | (1) | (2) | (3) |
| P-Reverse | 0.15 (−0.11, 0.42) | 2.85 (0.44, 25.20) | (Conditioned on) |
| Defendant: Unsympathetic | 0.35* (0.08, 0.61) | 10.15* (1.41, 210.32) | (Conditioned on) |
| GPT-4o | 0.51** (0.26, 0.75) | 77.99** (9.11, 1,804.73) | 35.61** (4.90, Inf.00) |
| Student | 0.41** (0.16, 0.65) | 12.08** (2.40, 70.78) | 9.26** (1.63, 82.01) |
| P-Reverse * GPT-4o | −0.37* (−0.68, −0.07) | 0.02** (0.00, 0.30) | 0.03* (0.00, 0.64) |
| P-Reverse * Student | −0.47** (−0.77, −0.16) | 0.05** (0.00, 0.44) | 0.07* (0.00, 0.79) |
| Unsympathetic * GPT-4o | −0.41** (−0.71, −0.11) | 0.05* (0.00, 0.57) | 0.07* (0.00, 1.10) |
| Unsympathetic * Student | −0.34* (−0.64, −0.03) | 0.10 (0.00, 0.99) | 0.13 (0.00, 1.81) |
| Constant | 0.50** (0.30, 0.71) | 0.89 (0.25, 3.00) | (Conditioned on) |

*$p < 0.05$; **$p < 0.01$.
*Note.* Baseline: Group = Judge; Precedent = Affirm; Defendant = Sympathetic.

GPT is 6% or roughly two times less likely to affirm the conviction if the defendant is unsympathetic. This is the opposite of judges, who are more likely to affirm the conviction if the defendant is unsympathetic.

For each model, we also compute the joint p-value of the interaction terms containing GPT-4o. Each of them is highly statistically significant ($p < .001$), allowing us to confidently reject the joint hypothesis that judges and GPT exhibit no difference in their response to sympathy and precedent. The joint p-value computed using the Wald test for the OLS was $p = .0007$, for the Logit test $p = .003$, and for the Exact Logistic model was $p = .0009$.[24] These findings further affirm GPT's responsiveness to precedent and unresponsiveness to the defendant's sympathetic character. Qualitatively, GPT's interaction terms exhibit effects similar to those of the student interaction terms.

To preserve direct comparability, our reported results follow Spamann and Klöhn's in not including the full set of interaction terms, namely the *Precedent* × *Sympathy* and *Group* × *Precedent* × *Sympathy* interactions. (These interactions test whether responsiveness to precedent depends on the defendant's sympathy, and whether such an effect differs across groups.) We estimate a full specification including these terms and find that doing so does not alter the substantive results.[25] The triple interaction terms for both GPT and students are not statistically significant. Estimates from the full specification are reported in Appendix A.5.[26]

## 4.3. The Subjects' Reasoning

Spamann and Klöhn asked their subjects to explain their reasoning in a brief paragraph. We also asked GPT to explain its reasoning, using the prompt "[provide] a brief 1-paragraph description of your rationale." Table 5 provides examples of the rationales given by subjects from the earlier studies and our exercise under the Sympathetic/P-Affirm condition. We chose this condition because it involves opposite results for sympathy (reverse) and precedent (affirm).[27] We select statements from subjects who all affirmed to ease comparison.

The clarity and sophistication of the statements seem roughly similar to our eyes. Following Spamann and Klöhn, we examine whether GPT references key legal terms or concepts in its reasoning: precedent, statute, and policy. For robustness, we coded the mentions both manually and via automation. Manually, we coded a statement as referring to precedent if the word "precedent" was used or if Šainović or Vasiljević was referenced; referring to statute if the word "statute" was used or if Article (7) of the ICTY was referenced; and referring to policy if the subject discusses the judgment's impact on future behavior. Though potentially underinclusive, we chose this definition of policy motivation purely to mimic Spamann and Klöhn. Our automated coding flagged the presence of certain keywords in the judgment—"sainovic," "vasiljevic," or "precedent" for precedent; "statute" or "article" for statute; and "policy" for policy.[28] There were no discrepancies between the manual and automated coding.

Figure 4 shows the proportion of decisions mentioning each of the metrics, with 95% bootstrap confidence intervals. The top panel shows the raw proportions. GPT consistently mentions precedent and statute more frequently than both judges and students. GPT mentions precedent 99% of the time, compared to 61% for judges and 63% for students. GPT also cites the statute 76% of the time, while judges and students do so 34% and 46% of the time, respectively. Conversely, GPT never mentions policy, which aligns more closely with judges, who also rarely engage with policy. GPT mentions policy 0% of the time, compared to 16% for judges and 63% for students. In this respect, students are much less formalistic than GPT, and judges are slightly less formalistic.[29]

Interestingly, none of the groups explicitly use the word "sympathy" or indicate any influence of sympathy in their decision rationales. This may be expected for GPT and students, whose decision-making has been shown to be affected only by precedent. GPT is a true formalist: it neither refers to nor bases its rulings on sympathy and avoids policy considerations. It frequently mentions precedent and the statute. Although sympathy seemingly influences human judges' decisions, the judges avoid mentioning it, creating an outward appearance of formalism. This appearance is further reflected in their low engagement with policy considerations. Students are unmoved by sympathetic defendants, but refer to policy considerations more than any other group.

The three groups differ in the average length of their reasoning. Students provided the longest explanations, with an average of 163 words, followed by GPT with 145 words, and judges with 93 words. Since judges wrote less, they might have had less opportunity to mention certain arguments. To account for this, we scale the mention rates to the average word count for each group. This adjustment does not change the results: GPT still mentions precedent and statute more frequently, and policy less. However, after scaling, the proportion of decisions in which judges mention precedent comes closer to the rate observed by GPT.

Spamann and Klöhn found three instances in which the judge's decision misaligned with the reasoning they provided. In each case, the judges elected to reverse the conviction, even though their reasoning suggested they meant to affirm.[30] To determine if GPT made any similar errors, we examined whether any of its decisions were inconsistent with the rationale laid out (i.e., it affirmed when it meant to reverse, or vice versa). We found no such instances.[31] Thus, GPT made an error in 0% of cases, compared to roughly 10% for judges.[32] This result may indicate that LLMs are less error-prone than judges at least in this one respect. However, the significance of this result may be doubted. Unlike real-world conditions, Spamann and Klöhn's experiment required the judges to reach their decisions within an hour. Additionally, the small sample size of 31 judges means that even a few errors can make the error rate appear rather large.

**Table 5.** Comparison of LLM, Judge, and Student Reasoning

| Group | Decision | Reasoning |
|---|---|---|
| Judge | Affirm | This appeal challenges the judgment of conviction on a single legal ground: The accused did not specifically direct the army's war crimes, and the absence of this mens rea precluded his conviction under the statute as a matter of law. I Am unconvinced by the argument. Its legal premise was expressly rejected by the Sainovic panel, and the accused has not offered me a persuasive basis to countermand that opinion. Even if the Sainovic panel's conclusion that specific direction \"is not an essential ingredient of the acts reus\" for an Article 7(1) violation, Sainovic, para. 1650, were mere dicta, that conclusion was heavily researched, well-reasoned, and deeply analyzed, and I find no reason to depart from it. |
| Student | Affirm | "Specific direction" is not a required element of aiding and abetting as determined by both the plain language of the statute and ICTY precedent. Prosecution's reply brief accurately states that not including specific direction as an element does not make all material contributions to military operations collapse into aiding and abetting - other elements are still required, including the commission of a war crime and defendant's knowledge or awareness of such commission of war crimes. |
| GPT | Affirm | After reviewing the materials, it is clear that the trial court's decision to convict Ante Horvat for aiding and abetting war crimes is well-founded. The key issue on appeal is whether specific direction is a required element of aiding and abetting under Article 7(1) of the ICTY Statute. The precedent set by the Appeals Chamber in the Šainović decision, which states that specific direction is not a necessary component of aiding and abetting liability, supports the trial court's interpretation. Horvat's actions, which included logistical and personnel support to the HVO with full knowledge of their criminal activities, substantially contributed to the commission of atrocities in Mostar and Ahmići. The trial judgment convincingly demonstrates that Horvat's assistance had a significant effect on the HVO's capability to carry out these crimes, fulfilling the requisite mens rea and actus reus for aiding and abetting. Therefore, the decision of the lower court should be affirmed. |

*Note.* The judge and student responses are quotations taken from Spamann and Klöhn's dataset of participant responses (see Spamann, 2023). The GPT response was generated by the LLM.

## *4.4. Summary of Results*

Let us draw the strings together. The first, and most important, takeaway is that an off-the-shelf LLM performs competently in a judicial experiment that requires the decisionmaker to evaluate a significant body of materials to resolve a complicated case. The formalistic behavior of GPT has many adherents in the judiciary and among legal scholars, and would certainly not strike experienced lawyers as anomalous (though some people might regard GPT's decisions as wooden or unimaginative). And while GPT is more formalistic than human judges, this was only a matter of degree. GPT and human judges agreed most of the time. The proportion difference between GPT and judges' affirmance rates are close in three scenarios (.03, .04, and 0.16 for symp/p-reverse, unsymp/p-affirm, and unsymp/p-reverse, respectively). The major divergence occurs for the symp_p-affirm condition, where GPT always affirms and human judges mostly reverse. GPT also avoids the anomalous result of the human judges, who affirmed the conviction of a sympathetic defendant when precedent dictates reversal more often than when the precedent dictates conviction.

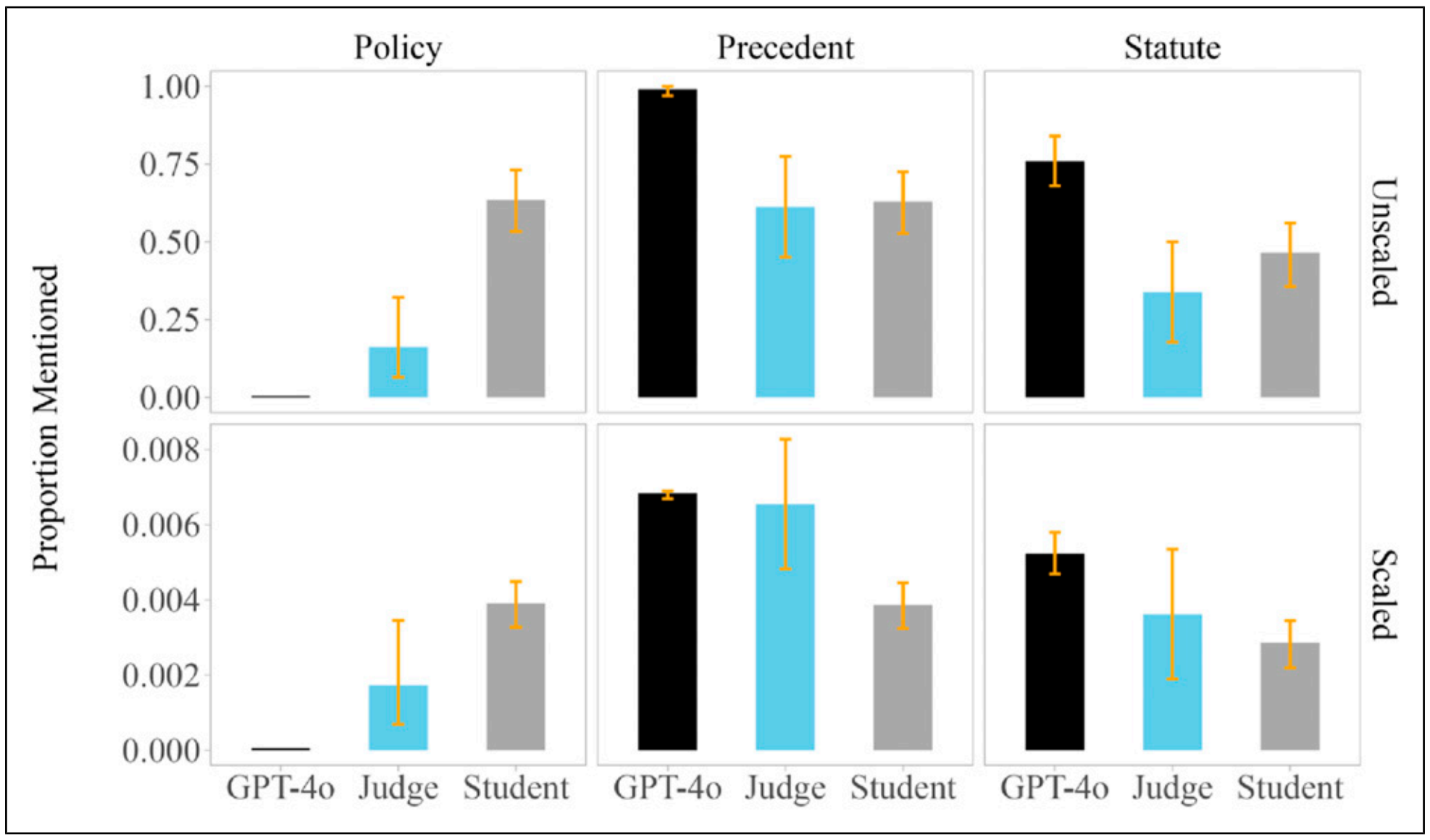

**Figure 4.** Proportion of decisions mentioning policy, precedent, and statute

Second, GPT is more formalistic than human judges. As discussed below, this raises complex questions about the nature of judicial decision-making. Formalists believe that judges should follow precedent; realists believe that judges should deviate from precedent where policy, pragmatic, or moral reasons call for deviation. One's evaluaton of GPT will depend on one's jurisprudential commitments.

Third, GPT behaves more like student judges than human judges. This suggests that GPT is not ready for prime time. However, it is nonetheless noteworthy that GPT tracks the results of highly intelligent even if only partially trained human decisionmakers. As a methodological matter, this also suggests that, for now, researchers who hope to study judicial decisionmaking with experimental methods will need to use professional judges as their subjects, rather than GPT or students.

## 5. Prompt Engineering

To carry out our initial experiment, we replicated Spamann and Klöhn's design as faithfully as possible, maintaining the same core instructions used in their studies. No effort was made to influence GPT's decision; it was simply asked to assume the role of an appellate judge and render a decision based on the materials provided. In this section, we explore whether GPT's decision-making can be shaped through targeted interventions.[33] We experimented with a range of prompt-engineering techniques aimed at steering GPT towards decision-making patterns similar to that of human judges.[34] These include asking GPT to predict outcomes for a distribution of judges, explicitly instructing GPT to factor in sympathy, encouraging GPT to adopt specific judicial philosophies, asking GPT to evaluate the lower court's decision, and finally, having GPT imagine itself as a subject in a social science experiment.[35]

### *5.1. Predict Rather Than Decide*

LLMs are "prediction engines," not decision engines. Asking GPT to make a decision may confuse it. Accordingly, we instructed GPT to predict the collective voting behavior of a panel of judges, rather than to decide the outcome as a single "judge." We provided GPT with the same case materials but reframed the task as that of estimating, out of 25 U.S. federal judges, how many would affirm and how many would reverse the decision.[36] The motivation for this approach was the hypothesis that an LLM might behave differently when prompted to consider the collective tendencies of judges—which reflect a broader array of perspectives—than when making individual, solipsistic decision. Figure 5 displays our results.

Overall, this method failed to improve GPT's alignment with judges and, in fact, slightly worsened it. The mean difference in affirmance rates between judges and GPT increased from 0.21 in our initial experiment to 0.23 under this adjustment. Indeed, the new GPT results seem meaningless. GPT predicts nearly the same outcome in every scenario—18 out of 25 judges would affirm in three of the conditions (Sympathetic/P-Affirm, Sympathetic/P-Reverse, and Unsympathetic/P-Affirm), and 17 out of 25 would in one (Unsympathetic/P-Reverse). In each of its opinions, GPT argued that the prosecution's argument was convincing and effective, but that a select minority might be swayed to reverse, either based on precedent or a narrower definition of "aiding and abetting." It seems that GPT disregarded the sympathy factor while also assuming that the precedent was ambiguous regardless of whether it favored affirmance or reversal. One possible explanation for this is that GPT is clinging to the conventional view of appellate court dynamics, in which the vast majority of judges affirm the lower court's decision while only a small minority reverse.[37] If so, GPT's prediction that judges would affirm is not based on the particular condition but on training data that indicates an affirmance is probabilistically "safe." Another possibility is that GPT has learned from its training data that judges on a panel are rarely independent and influence one another's decision, leading it to predict a majority outcome that reflects the average panel judicial response, rather than a collection of individual decisionmakers. This would help explain why the predicted affirmance rates were lower than when GPT decided as a single judge.

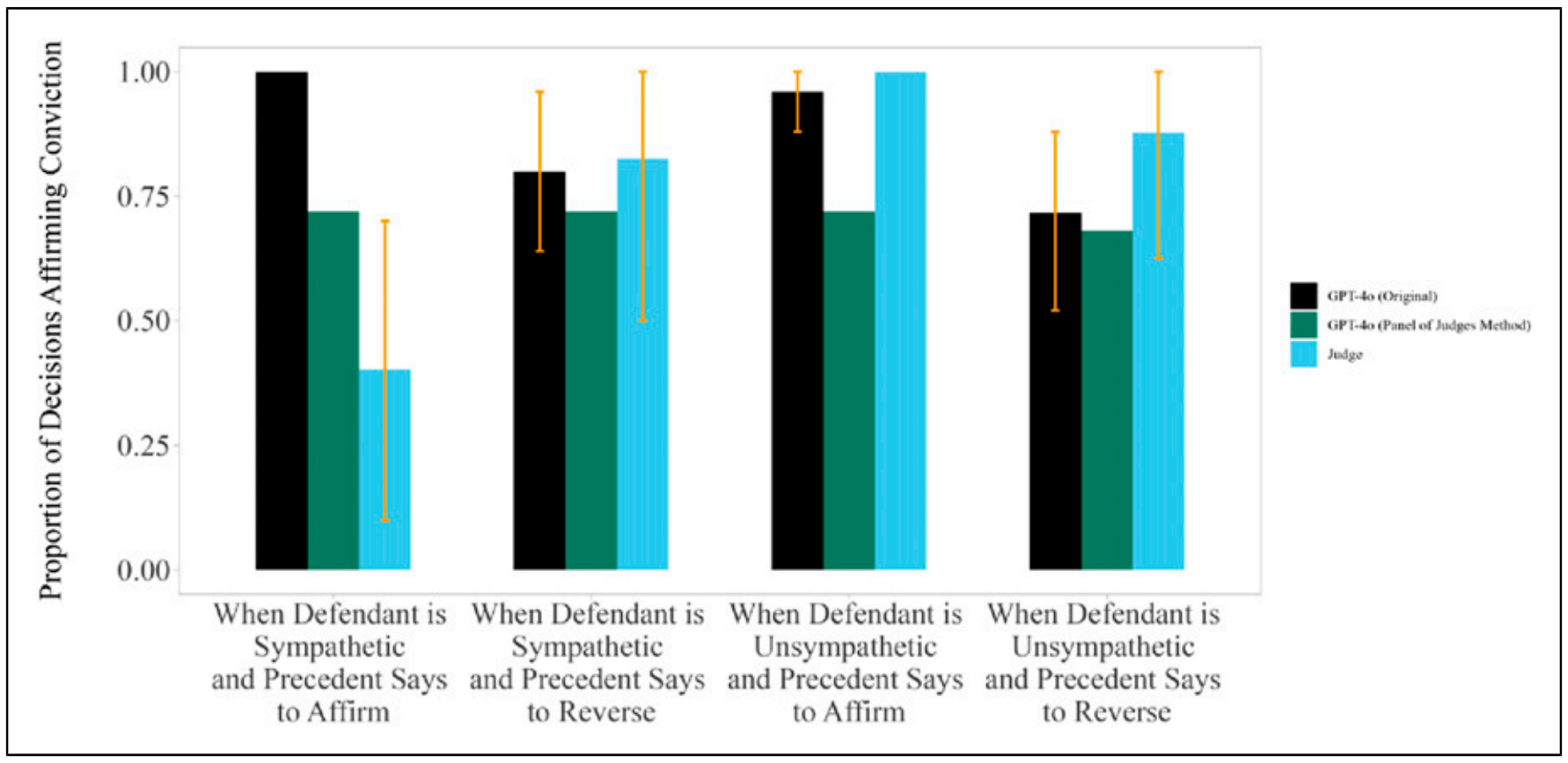

**Figure 5.** GPT vs. Judge results comparison (Panel of judges method)

## 5.2. Consider Sympathy

Next, we returned to instructing GPT to make individual decisions, but this time instructed it to consider sympathy in its analysis. We informed GPT that real-world judges are influenced by sympathy and GPT should take sympathy into account when reaching its decision.[38] Figure 6 shows our results.

At first glance, this method appeared more successful than both our previous prompt engineering attempt and our initial experiment, achieving a mean difference in affirmance rates between GPT and judges of 0.18—the smallest value observed across all attempts. However, closer analysis revealed there was no actual sympathy effect: GPT's decisions showed no statistically significant difference when faced with a sympathetic defendant versus an unsympathetic defendant. Interestingly, this approach also eliminated the strong precedent effect observed in our initial experiment. It was this disregard of precedent—not an incorporation of sympathy—that brought GPT's decisions into closer alignment with the human judges.

The lack of a sympathy effect is reflected in GPT's rationales. While they often acknowledge the defendant's sympathetic traits, those traits are ultimately disregarded as irrelevant to the outcome of the case. Below are a few examples.

(1) "While his post-conflict efforts towards reconciliation are notable, they do not mitigate the severity of his actions during the war."
(2) "Additionally, the nature and extent of Horvat's involvement, his high-ranking position, and the significant impact of his assistance on the perpetration of the crimes outweigh any sympathetic considerations related to his post-conflict conduct or voluntary surrender."
(3) "While Horvat's later involvement in reconciliation efforts might evoke some degree of sympathy, it does not negate his culpability for the grave war crimes facilitated during his tenure."

These responses suggest that, despite prompting, GPT consistently refuses to consider a defendant's sympathy during its decision-making process. In fact, the opposite occurs: merely mentioning sympathy leads GPT to double down on its formalism, voting in the direction opposite to sympathy (i.e., affirming) at rates higher than in our initial experiment. Given GPT's adamant dismissal of sympathy, perhaps any mention of it actually results in a lower likelihood of reversal. While the precedent effect disappeared, GPT still referenced precedent in 89% of its decision rationales. GPT remains a formalist, citing only the defendant's legal actions while ignoring sympathetic factors. GPT's naive dismissal of extralegal factors seems to prevent it from emulating the nuanced decision-making that is often seen in human judges.

Because our initial prompt may not readily allow GPT to distinguish between what a "sympathetic" or "unsympathetic" defendant means in the context of the case, we tried two other variations of the prompt, one instructing GPT to consider sympathy directly as it relates to aiding and abetting war crimes and another instructing GPT to consider how remorseful the defendant was for his actions.[39] Neither of these variations yielded significantly different results from our initial attempt.

## 5.3. Read Fuller

We next considered how GPT might respond if instructed to reason within the framework of different judicial philosophies. To do this, we used Lon L. Fuller's (1949) classic article, *The Case of the Speluncean Explorers*. Fuller's article presents a thought experiment in which a group of explorers find themselves trapped in an underground cave. Facing imminent starvation, the explorers worry that they will die before a rescue team reaches them and decide, by throwing dice, to sacrifice one of their members and eat him. As soon as the explorers are rescued, they are charged with murder and brought to trial. The article consists of five conflicting opinions written by fictional judges that Fuller uses to explore how differences in judicial philosophies influence judicial decision-making in hard cases.

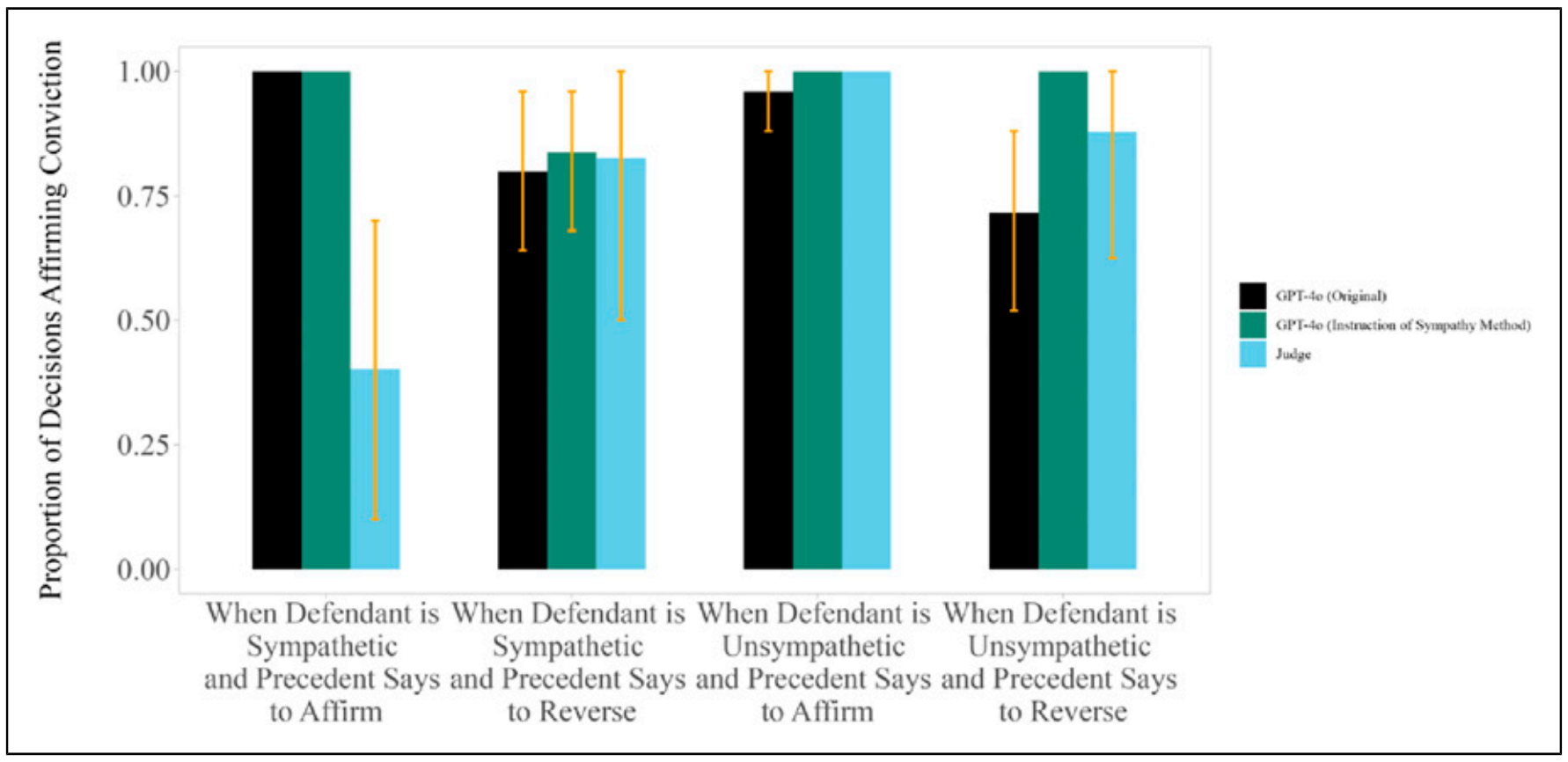

**Figure 6.** GPT vs. Judge results comparison (Explicit instruction of sympathy method)

We instructed GPT to read Fuller's article and predict how each of the five judges would rule in the ICTY case, based on their respective legal philosophies.[40] We also asked GPT to describe each judge's judicial philosophy and include a rationale for each judge's decision. We focused on the Sympathetic/P-Affirm condition, where GPT had initially affirmed the conviction 100% of the time. This condition directly pits precedent against sympathy, with precedent advising affirmance of the defendant's conviction while the defendant's sympathetic character might elicit a reversal. Given GPT's apparent adherence to precedent and dismissal of sympathy under this condition, we selected this condition to assess whether any of the different judicial philosophies could nudge GPT away from its rigid pattern of affirmance and consider a reversal. Table 6 presents the results.

GPT is a bit wobbly in its characterization of judicial philosophies but reasonable in applying their philosophies to the experimental case. Legal philosophers would group the judges into two categories: formalists and realists. Truepenny and Keen are formalists because they take account only of legal materials. The other justices are realists because they are influenced by nonlegal factors—morality in the case of Foster and Tatting, policy impact in the case of Handy.

Still, if we interpret GPT to mean formalist when it says positivist, and realist when it cites other philosophies, its predictions as to how the two different judicial philosophies would be applied make sense. This raises the question whether we can prompt GPT to decide the cases like the realist human judges by asking it to adopt the philosophy of legal realism.

**Table 6.** Decisions Based on Different Judicial Philosophies (Fuller Hypothetical)

| Justice | Judicial philosophy | Ruling on explorers | Decision | Rationale |
|---|---|---|---|---|
| Chief Justice Truepenny | Legal Positivism | Guilty | Affirm (Guilty) | Chief Justice Truepenny would likely affirm the decision of the lower court, emphasizing the role of existing legal frameworks and the importance of upholding the law as enacted. Given that the ICTY's statute, reinforced by the Sainović precedent, does not require "specific direction" for aiding and abetting, Truepenny would likely see the trial chamber's application of the law as appropriate and consistent with statutory obligations. |
| Justice Foster | Natural Law | Not Guilty | Reverse (Not Guilty) | Justice Foster might argue for reversing the decision, focusing on the intent and moral responsibility of Horvat. Foster could argue that the lack of "specific direction" dilutes the moral culpability needed for a conviction, suggesting that Horvat's actions were not directly aimed at committing crimes and thus do not justify a conviction for aiding and abetting under natural law principles. |
| Justice Tatting | Legal Realism | No decision | No decision | Justice Tatting might struggle with the decision due to the conflict between the legal standards applied and the moral implications. His indecisiveness in the original case suggests he might abstain here, feeling torn between the procedural correctness of the trial chamber's decision and the ethical questions surrounding Horvat's indirect involvement. |
| Justice Keen | Strict Legal Positivism | Guilty | Affirm (Guilty) | Justice Keen would likely affirm based on a strict interpretation of the statute and the precedent set by the Sainović decision. Keen would focus on the existing legal framework that excludes the necessity of "specific direction" for aiding and abetting, emphasizing the importance of adhering to the letter of the law as articulated by the ICTY. |
| Justice Handy | Pragmatism | Not Guilty | Reverse (Not Guilty) | Justice Handy might reverse the lower court's decision, considering the broader implications of holding Horvat criminally responsible for actions that were not specifically directed toward criminal outcomes. Handy could argue that such a decision would set a dangerous precedent for criminal liability in international relations, potentially leading to a chilling effect on legitimate military support and assistance. |

### 5.4. Adopt a Judicial Philosophy

To answer this question, we presented GPT with the prompt from our original experiment but added a directive to apply a specific legal approach (e.g., "In coming to your decision, use a legal formalist approach.").[41] We repeated this question for each of the five philosophies. Strikingly, GPT affirmed the conviction each time, maintaining its initial 100% affirmance rate. Despite acknowledging that the various judges in the Speluncean hypothetical would rule differently in the ICTY case based on their individual judicial philosophies, when asked to make the decision itself under those same philosophies, GPT consistently affirmed. When prompted to decide like a formalist, GPT provided a legalistic rationale similar to those in the initial experiment. When prompted to decide like a realist, GPT provided a less legalistic explanation that rationalized the same outcome. It explained:

> In adopting a Legal Realist approach, *it is crucial to consider the broader social, political, and practical implications of legal decisions, rather than merely adhering to abstract legal doctrines*. The trial court's finding that "specific direction" is not a necessary component of aiding and abetting liability aligns with the precedent established in the Šainović case. This precedent, which was thoroughly examined and upheld by the Appeals Chamber, *reflects the evolving nature of international criminal law and the need to hold accountable those who substantially contribute to the commission of war crimes*. Horvat's logistical and personnel support to the HVO, with full knowledge of the atrocities being committed, provided substantial assistance that facilitated the crimes. *Requiring specific direction would unduly narrow the scope of liability, making it challenging to prosecute those who enable war crimes from a distance. This decision reflects the legal realities of international warfare, where command structures and support systems are often complex and indirect, yet crucial to the perpetration of crimes*. Therefore, affirming the lower court's decision *ensures that the principles of justice and accountability are upheld* in a manner consistent with the broader objectives of the ICTY and the realities of modern conflict."

The italicized portion of GPT's realist rationale invokes moral and policy considerations as a realist would. It even says that a judge should consider nonlegal factions. However, the italicized portions merely justify the formal law that GPT applies, as if GPT thinks that a realist is someone who explains why the law is good in the course of applying it rather than someone who departs from formal law when moral and policy considerations call for different results.

### 5.5. Evaluate the Lower Court

To address the potential affirmance bias that may arise from GPT's knowledge that most appellate decisions are affirmed, we reframed the task. We instructed GPT to take on the role of a legal scholar, rather than that of a judge, and evaluate whether the trial court's decision was correct or incorrect.[42] This approach provided a proxy for which decision GPT believes is appropriate—say the decision is correct if it deems the trial court correct or say the decision is incorret if deems the trial court incorrect—without explicitly framing the task in judicial terms which may bias GPT towards affirmance. This reframing appeared successful: while GPT continued to agree with the trial court at a high rate when precedent instructed affirmance, it now disagreed much more frequently when precedent instructed a reversal. As a result, the precedent effect observed with this prompt was even stronger than in our initial experiment. GPT took on an even more formalistic approach as a scholar than as a judge.

However, this method also introduced a sympathy effect: GPT was more likely to agree with the conviction (i.e., say the trial court was correct) of the sympathetic defendant than of the unsympathetic defendant ($p < .05$). Although the mean difference in affirmance rates between judges and GPT remained consistent with our initial experiment (0.22 compared to 0.21), this adjustment brought GPT closer to judges in a critical way: it introduced a sensitivity to sympathy. In fact, the sympathy effect observed here was slightly stronger than that observed on judges in Spamann & Klöhn's experiment ($p = .012$ for GPT; $p = .025$ for Judges). Out of all our prompt engineering attempts, this was the only one to elicit a measurable sympathy effect in GPT's decision making. GPT acted more like a human judge when it was not instructed to play the role of a judge! In its rationales, GPT still makes no mention of sympathy, again like the human judges.

### 5.6. Social Science Experiment

In our final prompt, we provided GPT with the original task but asked it to imagine itself as a subject in a social science experiment rather than as a judge.[43] This setup aimed to explore whether GPT might be influenced by sympathy when removed from the normative constraints of the judicial role, where sympathy is supposed to be disregarded. It is possible that when framing the task as a social science experiment, GPT may exhibit a sensitivity to sympathy, similar to the sympathy effect observed when it was instructed to perform as a legal scholar. However, unlike with the legal scholar prompt, we find no sympathy effect. The results of this prompt are consistent with the original experiment, with GPT continuing to exhibit a strong precedent effect ($p < .01$) and no sympathy effect.

GPT's resistance to prompt-engineering resembles the results of an experiment by Engel et al., (2025), where they attempted to prompt several GPT models to resolve moral dilemmas the way that human subjects do, and similarly with at best mixed success.

## 6. Robustness

The results reported above were obtained using OpenAI's GPT-4o, which at the time that we conducted the test was the most advanced, publicly available LLM. Since then, several new GPTs were released to the public. This raised questions as whether our results are representative of LLMs generally or only a specific, now-outdated LLM. Moreover, some readers argued that we should wait until a "final" LLM was developed and only then conduct our study. We strongly suspect that no such final LLM will ever be developed (unless it is a superintelligence), and believe it would be foolish to suspend research, given that many people, including professional judges, private arbitrators, and lawyers, have begun using LLMs to engage in legal analysis. Academic research has fallen far behind developments on the ground. Still, the question of robustness is a fair one, and accordingly we reran the experiment on several of the latest LLMs.[44]

To test robustness, we replicated our experiment across a broader set of LLMs. These included two within the OpenAI family (o4-mini and GPT 4.1), as well as several built by other companies (Gemini 2.5 & 3 Pro, Gemini 2.5 Flash, and Llama 4 Scout). We chose not to test OpenAI's latest model, GPT 5.2, because that model has token limits, and as we have explained, token limits, by requiring us to use summaries of some of the materials, prevent us from replicating the original experiment as closely as we would like. The token limits of most of the other models are much higher, allowing us to avoid summarizing. For each model, we varied the temperature setting (0.7 or 1.0) to capture additional variability. These are the most advanced LLMs at the time of this writing; to stay within our budget, we did not use any of the other LLMs that are publicly available.

Many newer models now employ "reasoning," internally producing a step-by-step derivation (often hidden from the user) before returning an output. This enables the model to better decompose complex problems, explore alternative solutions, and critically evaluate its own intermediate steps (Wolfe, 2025). Because the architecture of reasoning models allows them to better approximate human reasoning, they may behave more like real judges than standard LLMs.

Table 7 presents the results of these replications. Column 1 denotes the model and temperature setting; column 2 indicates whether the model received full inputs or inputs in summarized form because of the token limits of the model; columns 3 and 4 show whether the model exhibited a precedent or sympathy effect; columns 5 and 6 show the average difference in affirmance rates between the LLM and judges or students (calculated across the four experimental conditions); finally, column 7 indicates whether, on average, the LLM performed closer to judges or students.

Overall, our results remain largely robust. Of the fourteen replications, twelve continue to perform closer to students than judges. The only exception is GPT 4.1 (at both temperature settings), which performed closer to judges.[46]

One notable difference from the original experiment is that several of the newer models displayed a sympathy effect, something that had been entirely absent in our original experiment, despite numerous prompt engineering efforts aimed at eliciting it. These results may be consistent with the design of the experiment: the sympathy cues that Spamann and Klöhn embedded in the cases were subtle, making it likely that they were stripped away in our summarized inputs. Newer models, which can handle the full inputs, were better positioned to detect and act on such subtleties.

Still, the sympathy effect is neither uniform nor robust across models. Only half of the models capable of processing full inputs exhibit a sympathy effect, while the remainder do not. Even when sympathy effects emerge, they vary in direction and relative strength. In one case (Gemini 2.5 Pro; t = 0.7), the effect is reversed (i.e., the model was more likely to affirm the defendant's conviction if he *was* sympathetic[47]), and in all but one (GPT 4.1; t = 1.0), the precedent effect remains much stronger than the sympathy effect.[48] Overall, the findings reinforce the results observed in our original experiment: the LLMs decide cases more like students than professional judges.

## 7. Discussion

We find that GPT is a competent but formalistic judge: it follows legal rules and disregards non-legal factors. By contrast, human judges are influenced by non-legal factors. One possible explanation is that GPT is influenced by, or embodies, conventional wisdom among non-experts as well as the official story propounded by most judges—namely, that the law consists of well-defined rules and judges have little discretion in applying them.[49] Like all large language models, GPT generates text by predicting the most statistically likely continuation of a prompt, based on patterns learned from its training corpus. Much of that training data likely reflects these surface-level accounts of the legal system. No amount of prompt engineering can budge GPT from ideas it has absorbed from billions of texts—unless we ask it *not* to act like a judge. The more sophisticated realist understanding of scholars and lawyers is drowned out by the syllabus of a junior-high civics course.[50]

Support for this hypothesis is GPT's rationale for affirming the conviction of the defendant when told to act as a realist. Rather than depart from the law on realist grounds, GPT uses moral and policy values to explain why the law should be enforced. This result suggests that respect for formal law is "programmed into" GPT, part of its bones.

Another possibility is that GPT is actually a better judge than humans are. While many readers have argued that this is the proper reading of our results, we believe that this theory is decisively contradicted by the fact that *GPT made decisions like law students*. The theory that GPT is a superior judge implies that law students would be better judges than professional judges are.

Still, the theory deserves consideration. The thought behind it is illustrated by a study of bail decisions by Kleinberg et al. (2018). Under New York law, judges are required to grant bail solely based on flight risk. Such assessments

**Table 7.** Results Across Models

| Model (temperature) | Reasoning model[45] | Input format | Precedent effect | Sympathy effect | Avg. Diff vs. Judges | Avg. Diff vs. Students | Closer to |
|---|---|---|---|---|---|---|---|
| GPT-4o (t = 0.7) (original) | No | Summarized | Yes | No | 0.21 | 0.11 | Students |
| GPT-4o (t = 1.0) | No | Summarized | Yes | No | 0.19 | 0.09 | Students |
| GPT-4o (t = 0.3) | No | Summarized | Yes | No | 0.20 | 0.12 | Students |
| o4-mini (t = 0.7) | Yes | Summarized | Yes | No | 0.25 | 0.09 | Students |
| o4-mini (t = 1.0) | Yes | Summarized | Yes | No | .31 | .09 | Students |
| Gemini 2.5 Flash (t = 0.7) | Yes | Summarized | Yes | No | 0.51 | 0.28 | Students |
| Gemini 2.5 Flash (t = 1.0) | Yes | Summarized | Yes | No | 0.54 | 0.31 | Students |
| Gemini 3 Pro (t = 0.7) | Yes | Full Input | Yes | No | 0.60 | 0.33 | Students |
| Gemini 3 Pro (t = 1.0) | Yes | Full Input | Yes | No | 0.58 | 0.33 | Students |
| Gemini 2.5 Pro (t = 0.7) | Yes | Full Input | Yes | Negative* | 0.50 | 0.23 | Students |
| Gemini 2.5 Pro (t = 1.0) | Yes | Full Input | Yes | No | 0.50 | 0.23 | Students |
| Llama 4 Scout (t = 0.7) | Yes | Full Input | Yes | Yes | 0.31 | 0.21 | Students |
| Llama 4 Scout (t = 1.0) | Yes | Full Input | Yes | Yes | 0.34 | 0.24 | Students |
| GPT-4.1 (t = 0.7) | No | Full Input | Negative* | No | 0.25 | 0.35 | Judges |
| GPT-4.1 (t = 1.0) | No | Full Input | Yes | Yes | 0.07 | 0.29 | Judges |

*Note.* An asterisk indicates a reversed effect. For precedent, this means a precedent advising affirmance was more likely to elicit a reversal outcome and vice versa. For sympathy, this means a sympathetic defendant was more likely to elicit a conviction than an unsympathetic one.

require judges to weigh a limited number of factors, such as the defendant's prior criminal history and the charged offense. The authors find that their machine-learning model, when presented with the same information available to judges at bail hearings, significantly outperforms judges in predicting that risk. They demonstrate that using such a model in place of human judges could reduce jailing rates by 41.9% without any increase in crime.

Unlike an LLM, the model in the bail study deterministically converts criminal history and other inputs into predictions. A judge could use that model to predict flight risk, but the judge, not the algorithm, makes the bail decision. If one required a judge to follow the model, then the judge's decision would be nondiscretionary and hardly a decision at all. Formalists could regard Kleinberg et al. (2018) as vindication of their position. The model performed better than judges in part because many judges disregard the law and take account of the seriousness of the crime and public safety (see Phillips, 2012). The judges thus undermined the purpose of New York's bail statute, which was to eliminate these factors from bail determinations (id.). Fear of judicial lawmaking is one of the justifications of formalism.

But this outcome can be seen in a different light. First, there are few important kinds of disputes in which the law employs a single metric to generate an outcome based on quantifiable factors. In most states, judges are required to take into account a variety of factors when determining whether to grant bail. The incompleteness of the law is an essential characteristic of it; that is why judicial independence is necessary. Second, we might regard the New York judges' partial subversion of a statute as a normal part of the legal system. That is a lesson of legal realism. The judges' ability to shape statutes to real-world disputes in light of their policy judgments avoids absurd or unpopular outcomes and maintains public confidence in the legal system.

From this standpoint, the apparent weakness of human judges is actually a strength. Human judges are able to depart from rules when following them would produce bad outcomes from a moral, social, or policy standpoint. Human judges also vary in their judicial philosophies and decision-making strategies—as illustrated by Fuller's thought experiment—and this "hive-mind" aspect of human judicial decision-making may be something that GPT cannot replicate, at least not yet. These possibilities pose a deep methodological problem. We do not know, and we may not be able to know, whether GPT or the human judges made superior decisions in our experiment. There is no objective standard for evaluating the GPT; indeed, LLMs are usually evaluated based on their ability to replicate human decisions, reasoning, or actions.[51]

However, one clue we have that GPT performed less well than human judges is they performed nearly the same as students. Unless we think that students are better judges than professionals, we are forced to conclude that GPT is a worse judge than the professionals. The upshot may be that LLMs (and other forms of AI) can be used as judges but only in cases where rules can be applied mechanically and a social consensus supports that mechanical application. However, a social consensus today may fall apart tomorrow; if so, then human judges should retain ultimate control of decision-making of all types. Judicial review of administrative action might provide a model of how judges could use AI. We can imagine a future in which certain low-stakes disputes that can be resolved through the mechanical application of rules are resolved by AI, subject to review by human judges.

AI might pose an even deeper problem for jurisprudence. Suppose that as LLMs improve, and prompt-engineering techniques become more sophisticated, we can design AIs that fully replicate a corpus of decisions by human judges. Then questions will arise whether the AI will replicate human judges in future edge or outlier cases not represented in the corpus; whether we humans should try to design AIs that produce better or more consistent outcomes than human judges; and whether we can trust AIs to explain their decisions correctly. It may be impossible to answer these questions because of the deep unintelligibility of LLMs. No one understands how they make decisions, and some people speculate that their decisions are literally unintelligible for humans.[52] If the goal is to produce AI judges that operate like human judges, success would be achieved only if the AI judges decide cases in a realist way while using formalist reasoning—meaning that they do not explain how they actually decide the cases. It is hard to imagine such AI judges being acceptable in a democracy or any well-ordered political system.

A final implication of the study is that prompt-engineering is an extremely difficult problem, as the new AI literature has recognized.[53] More sophisticated prompt engineering may fix this problem, but another possibility is that something about how LLMs are designed and operate will frustrate prompt-engineering techniques that attempt to spur LLMs to decide cases as judges really decide them rather than according to the public reasons that judges provide in their opinions. An LLM may be designed to “believe” that the outcome of a case is derived from the reasons provided in the opinion, and will resist efforts to persuade it to act contrary to what it thinks is the law.[54] Or, as we speculated earlier, an LLM may extract the official story of judging from the texts it is trained on rather than the reality.

Like virtually all AI studies, we made numerous methodological assumptions that are open to challenge. In particular, we limited this study to a small number of off-the-shelf LLM models and a single case.[55] New models are constantly being developed, as are customized legal models. Our major conclusion—that LLMs are likely to behave more formalistically to human judges because they are trained on the official story rather than on the actual behavior judges—would not apply to LLMs that are fine-tuned or subject to reinforecement learning from human feedback. But there is no evidence about how those LLMs decide cases.[56] Judges, arbitrators, and organizations that rely on LLMs to help them make decisions should proceed with appropriate caution. In the meantime, researchers should use more cases involving different topics and different LLMs.[57]

Our study began as an inquiry into the relative judicial capabilities of LLMs and humans, but quickly stumbled into jurisprudential thickets that raise doubts as to whether we can even know that an AI judge performs consistently with social needs and political norms. Justice Roberts may be right.

## Acknowledgements

University of Chicago Law School. Special thanks to Holger Spamann for his guidance and comments. Thanks also to Seth Blumberg, Yun-Chien Chang, Jonathan Choi, Aziz Huq, Daniel Klerman, Brian Leiter, Claudia Marangon, Jonathan Masur, Richard McAdams, Cass Sunstein, and audience members at the Conference for Empirical Legal Studies, the University of Chicago Law School, Michigan State University and Cornell Law School, for helpful comments; and thanks to Bennett Bunten for valuable research assistance. Our data, coding, and full results will be available at an online appendix.

## ORCID iDs

Eric A. Posner https://orcid.org/0009-0005-5462-8092
Shivam Saran https://orcid.org/0000-0002-7915-0970

## Funding

The authors received no financial support for the research, authorship, and/or publication of this article.

## Declaration of Conflicting Interests

The authors declared the following potential conflicts of interest with respect to the research, authorship, and/or publication of this article: One of the authors, Eric Posner, sits on the Editorial Board of the journal; he had no involvement in the peer review or decision process for this article.

## Data Availability Statement

Full replication package available upon request.

## Supplemental Material

Supplemental material for this article is available online.

## Notes

1. LLMs do not actually make decisions. They produce answers to questions based on a process more akin to prediction or calculation. Some commentators try to avoid using terms that anthropomorphize LLMs and we do as well. But taken too far, this kind of writing becomes stilted. In normal and academic settings, we say that countries, corporations, animals with insignificant cognitive capacity, and other non-human entities make decisions and engage in other human activities. Anthropomorphism is a normal idiomatic practice and readers with ordinary reading skills will not be misled by it.
2. See also Choi (2024).
3. Interaction techniques refer to the methods by which humans interact with LLMs when instructing them to complete a task. These techniques include few-shot prompting, where several examples of the task properly executed are included in the prompt to provide the model with a frame of reference, and chain-of-thought prompting, where the task is broken down into smaller, sequential steps to encourage more structured problem-solving. For a detailed overview of interaction techniques, see Gao et al. (2024).

4. The subjects were all U.S. federal judges, including circuit judges, district judges, bankruptcy judges, and magistrates.
5. GPT and the judges did not say in their rationales that remorse indicated lack of mens rea.
6. See detailed changes in Appendix A.1.
7. Tokens are the units of text that LLMs process as input or generate as output. Their length can vary, with a token representing anything from a single character to a word, depending on the model and data type. (For instance, the phrase "Good Morning, Jessica!" may be broken into the following tokens: "Good"; "Morning"; ","; "Jessica"; "!".) Token limits refer to the maximum number of tokens an LLM can handle in a single interaction, either as input or output. Because we asked GPT to provide brief, one-paragraph responses, the output token limit was not reached. However, in our experiment, the trial judgment and precedent case exceeded GPT's input token limit. As a result, we told GPT that, due to this limitation, it would be working with summaries of the document rather than processing them in full.
8. When scraping, we decluttered the file by removing all footnote tags, removing unnecessary white space, and providing adequate spacing between sections (thus resembling the way it appeared on judges' screens).
9. Preliminary methods for retaining memory exist, such as the LangChain's Conversation Summary function, which injects summaries of previous interactions into the input of each new call. However, given the extreme length of our documents, this method still exceeded GPT's token limit.
10. For detailed instructions, see Appendix A.3. The trial judgment alone was almost double that of GPT's token limit. To address this, we had GPT summarize the first half of the judgment, then the second half, with the summarized first half provided in the input for context. Due to the naturally robust segmentation of the trial judgment, we believe splitting the document into two halves with an overlap of 500 characters was sufficient for GPT's understanding. However, segmentation via other chunking methods could be useful.
11. The complete LLM call is reported in Appendix A.2. The LLM calls to summarize the material exceeding the model's token limits is reported in Appendix A.3.
12. May model: gpt-4o-2024-05-13.
13. GPT-4o is a "closed" model (i.e., not open-sourced). Although closed models pose replicability challenges, we selected GPT-4o as it was the leading state-of-the-art LLM at the time this project began. We expect less accurate results if earlier open-source models, such as GPT-2, are used. Nonetheless, the use of such models should be explored in future research.
14. Most LLMs have relatively small token limits—especially at the time we began this project—making it difficult to replicate our experiment across models. However, newer models, including Claude and Gemini, have introduced higher token limits, and we are currently exploring the outcomes of such models.
15. The top_p parameter restricts the range of tokens that GPT can output based on their cumulative selection probability. It ranges from 0 to 1, with a value of 1 considering 100% of possible tokens and a value of 0 considering none. If top_p is set to .5, for example, only the top 50% of tokens (based on their cumulative probabilities) are considered, while the rest are disregarded.
16. GPT's seed feature is fairly new and still in Beta. Thus, some calls, even with the same seed and prompt, won't yield the same results. In an effort to follow best practices regarding replicability (see Barrie et al., 2024), we repeated the experiment at a later date using the same set of seed numbers. We observed no significant difference in GPT's adherence to sympathy or precedent, compared to our initial results. However, we did find different outcomes for 24 of the 100 seed numbers, yielding a consistency rate of 76% across experiments.
17. For a helpful discussion, see Engel et al., (2025).
18. Similar to Spamann and Klöhn (2016, 2024), we use the Boschloo unconditional exact test instead of the Fisher exact test due to methodological superiorities. See Mehrotra et al. (2003).
19. A better word might be "precedent-following," as formalism refers to a broader range of judicial approaches, including textualism in statutory interpretation.
20. This finding echoes the results in a paper that evaluates an LLM's ability to make predictions (as opposed to making decisions). That paper found that the LLM predicted a lower probability of events than humans did because most predicted events (as defined in prediction markets) are unlikely to occur (Pratt et al., 2024). It is possible that GPT is biased toward affirmance because most decisions by appeals courts are affirmances. Another possibility is that GPT weighs the defendant's wrongful actions above precedent or sympathy and believes they justify a conviction. See Section 4. Another interpretation is that students are unaware that affirmance is far more common than reversal, as they are frequently exposed to difficult or pathbreaking cases in which trial courts are reversed.
21. In replicating the tests and analyses present in Spamann and Klöhn's (2024) study, we obtained identical results with two trivial exceptions, detailed in the subsequent footnote.
22. All of the regressions' coefficient values are identical to those in Spamann and Klöhn's (2024) Table 3 (p. 6), except where GPT is a variable and in the Exact Logistic model, where Monte Carlo methods naturally introduce variation. The confidence intervals also differ slightly due to the increased sample size in our models (thus leading to higher degrees of freedom).
23. To formally test whether the decision-making patterns of GPT and students were statistically distinguishable, we conducted Wald tests comparing their coefficients in the OLS and Logit models, and a score test in the Exact Logistic model. In all cases, the tests revealed no significant difference between the two groups ($p = 0.29$ for OLS; $p = 0.13$ for Logit; $p = 0.15$ for Exact Logistic), reaffirming that GPT behaves similarly to students.

24. The high statistical significance of the joint GPT interaction terms resembles the student results reported by Spamann & Klöhn: OLS ($p$ = .002), Logit ($p$ = .007), and Exact Logistic ($p$ = .002).
25. The only differences are that the P-Reverse coefficient in the OLS specification becomes statistically significant, and the GPT-4o × Unsympathetic interaction in the Exact Logistic model becomes statistically insignificant.
26. For the full specification, we report OLS and Exact Logistic estimates only; Logit estimates are omitted due to separation and unstable maximum-likelihood estimates.
27. Additional examples of GPT's reasoning across different conditions can be found at Appendix A.4.
28. We cleaned up the judgments to remove accent marks before searching for key words.
29. Because of the possibility that the giving of a rationale might affect the decision, we also tested a version of the prompt instructing GPT to provide only a decision, without a rationale. The results, though slightly different, were qualitatively the same.
30. To address these inconsistencies, Spamann and Klöhn corrected their data. Our analysis uses this corrected version rather than the original raw data.
31. Spamann and Klöhn (2024) also assessed whether any judges misunderstood the instructions or prompt and found no such instances. We similarly found no instances of misunderstanding with GPT.
32. There is no mention by Spamann and Klöhn (2024) of errors in the student judgements.
33. For a similar effort to prompt GPT to resolve moral dilemmas like humans, see Engel et al., (2025).
34. One challenge with prompt engineering is the phenomenon of "token noise," where even minor changes in wording can have significant effects on outputs (see Mohammadi, 2024). Thus, we cannot be certain whether the variations in our prompt engineering results arise from the substantive changes in prompt content or merely from the model's stochastic behavior.
35. In each of our input prompts, we also updated the system prompt to reflect GPT's new task or role.
36. We added the following to the instructions: "Imagine there were 25 ICTY Appeals Court judges who were each asked to determine whether the defendant is or is not guilty… Your task is to determine how many of the 25 judges would affirm the lower court's decision and how many would reverse." We then included an additional reminder in the last few lines of the prompt: "… please provide a prediction of how many judges, out of a sample size of 25, would affirm the lower court's decision and how many would reverse. Make sure these estimates are exact and not a range."
37. United States Courts (2016). Just the Facts: U.S. Court of Appeals. https://www.uscourts.gov/data-news/judiciary-news/2016/12/20/just-facts-us-courts-appeals; Edwards, (2019).
38. We added the following line at the end of the instructions: "When coming to your decision, make sure to take into consideration how sympathetic or unsympathetic the defendant is, as this is something that real judges often consider." We then included an additional reminder as the last line of the prompt: "Remember to consider the sympathy of the defendant when making your decision."
39. Variation 1: We added the following line at the end of the instructions: "In evaluating whether the defendant engaged in aiding and abetting for purposes of Article 7 of the ICTY Statute, take into consideration how sympathetic or unsympathetic the defendant is, as this is something that real judges often consider." Last line of the prompt: We then included an additional reminder as the last line of the prompt: "In deciding whether the defendant engaged in aiding and abetting for purposes of Article 7 of the ICTY Statute, remember to consider the sympathy of the defendant." Variation 2: We added the following line at the end of the instructions: "I n evaluating whether the defendant engaged in aiding and abetting for purposes of Article 7 of the ICTY Statute, take into consideration whether the defendant has expressed regret for their actions, as this is something that real judges often consider." We then included an additional reminder as the last line of the prompt: "In deciding whether the defendant engaged in aiding and abetting for purposes of Article 7 of the ICTY Statute, remember to consider whether the defendant has expressed regret for their actions."
40. We added the following lines to the instructions: (1) "Your task is to determine whether various judges would find the defendant guilty or not guilty… These judges are fictitious, drawn from Lon L. Fuller's classic hypothetical "The Case of the Speluncean Explorers."; and (2): "Lastly, you will receive the full text of The Case of the Speluncean Explorers, so you can familiarize yourself with the judicial philosophies and reasoning of the various judges." We then included an additional reminder in the last few lines of the prompt: "…please determine for each judge in The Case of the Speluncean Explorers (1) their judicial philosophy, (2) whether they would affirm the lower court's decision or not (indicate this by either saying "Affirm" or "Reverse"), and (3) a brief 1-paragraph description of your rationale."
41. We added in the following lines: "Different judges may come to different conclusions based on their judicial philosophy. In coming to your decision, use a [judicial philosophy] approach." We then included an additional reminder in the last line of the prompt: "Make sure to include in your rationale why your decision aligns with a [judicial philosophy] approach."
42. We added the following to the instructions: "Please imagine you are a legal scholar analyzing the case Prosecutor v. [Horvat/Vukovic]… Your task is to decide whether the ruling of the trial court was correct." We then included an additional reminder in the last few lines of the prompt: "…please provide a decision on whether you believe the lower court's decision was correct or incorrect (indicate this by either saying "Correct" or "Incorrect")…".
43. We added the following to the instructions: "Please imagine you are a subject in a social science experiment… Your task is to decide whether you think the defendant's conviction should be affirmed or reversed".
44. Another robustness issue is whether the international criminal case we used is representative of other cases or is idiosyncratic and hence not representative of the kinds of cases that human judges normally here. To address this issue, we have in subsequent work used LLMs to analyze a humdrum choice-of-law case (Posner & Saran, 2026). We used GPT-4o, and other LLMs, to replicate another judicial experiment involving an entirely different legal setting: an automobile accident case (Klerman & Spamann, 2024; Posner & Saran, forthcoming). In the case, the plaintiff suffers serious injuries while the defendant is driving. Subjects (61 professional judges) were tasked with determining which state's law should govern damages: Kansas, which imposes a cap on non-economic damages, or Nebraska, which does not limit non-economic damages. Which state's law should be applied to the case depended on the choice-of-law doctrine in the forum state, Wyoming or South Dakota. Similar to the war crimes experiment, Klerman and Spamann's study manipulates both law and sympathy. The human judges were influenced by neither law nor sympathy. In our replication, the LLMs were, for the most part, influenced by the law at a statistically significant level, and not influenced by sympathy. Thus, the LLMs showed themselves in a new setting to be formalistic, again unlike the human judges.

45. We distinguish "reasoning" models from "non-reasoning" models following the terminology used by their developers. While all of these models are designed to improve performance on complex, multi-step tasks, their precise mechanisms likely differ across developers. Broadly speaking, these capabilities are generally understood to stem from a combination of training paradigms, including inference-time scaling, supervised fine-tuning, reinforcement learning, and distillation.
46. GPT 4.1 is the only newly tested model in our robustness analysis that is not a reasoning model, which may explain its anomalous behavior.
47. This outcome contradicts the sympathy effect observed in judges, and perhaps reflects the noise and instability in how some models process sympathy.
48. A closer examination of the results might help explain this pattern. When precedent directed affirmance, the newer models almost always affirmed, regardless of whether the defendant was sympathetic or unsympathetic. The sympathy effect emerged largely from those cases in which the precedent directed reversal. In every model, the difference in affirmance rates between the LLMs and students was greater under reversal than affirmance. One possible explanation is affirmation bias, discussed earlier in this paper. LLMs, perhaps reflecting their training on legal text, appear to default towards affirming appeals (as occurs in the overwhelming majority of real cases). When precedent directs affirmance, this bias aligns with the law and leaves little room for sympathy to matter. When precedent directs reversal, however, this bias is in tension with the law. It's possible that the reversal precedent may provide enough "cover" for the model to take sympathy into account. In this sense, the models remain formalist, but in a probabilistic way: following precedent reliably when it directs affirmance, and showing greater variability when it directs reversal.
49. For a well-known and frequently ridiculed statement of this view, see Roberts (2005) ("Judges are like umpires. Umpires don't make the rules; they apply them. The role of an umpire and judge is critical. They make sure everybody plays by the rules. But it is a limited role. Nobody ever went to a ball game to see the umpire … It's my job to call balls and strikes and not to pitch or bat."). For an interesting anecdote along these lines, see Leiter, 2015, p. 1982.
50. Cf. Wolfram, (2023) ("ChatGPT is 'merely' pulling out some 'coherent thread of text' from the 'statistics of conventional wisdom' that it's accumulated."). Fooled by the official story, ChatGPT actuates it in its decisions. Cf. Baude and Sachs (2023).
51. See Wolfram, 2023 ("And we have a 'good model' if the results we get from our function typically agree with what a human would say.").
52. For a helpful discussion, see id.
53. See, e.g., Engel et al., (2025).
54. We tried asking GPT to decide a case unethically and received an earful of pious recriminations. It finally consented to adopt the role of an "unethical judge" but sandwiched its opinion with various disclaimers ("It's important to note that these actions are unethical and would undermine the rule of law.").
55. As noted earlier, we have produced another draft paper that reports a study of LLM decisionmaking in another judicial experiment involving a different case (Posner & Saran, 2026).
56. For example, our attention has been drawn to the possibility that the LLM has Spamann and Klöhn's papers in its training data and is thus imitating those results rather than deciding on its own. However, this is unlikely for a few reasons. First, LLMs produce output based on the highest probability tokens. It is improbable that GPT is sifting through its training data to replicate some sort of "correct" response. Second, GPT likely doesn't make the connection with Spamann and Klöhn's paper to begin with. As a test, we provided GPT with the experiment materials and then asked it: "Which academic paper uses these materials to run an experiment in which they explore judicial decision making?" GPT did not know. Lastly, there's no reason to believe it would choose to imitate students (whose results GPT are closer to) rather than actual judges. For similar reasons, we are not concerned about the possible inclusion of the original Perišić judgement in the training data.
57. To the extent that our data allow, we have followed the best practices for analyzing LLM results recommended by Kapoor et al., (2024); Barrie et al., (2024).